\documentclass[12pt]{article}
\usepackage[utf8]{inputenc}
\usepackage[T1]{fontenc}

\usepackage[compress]{cite}
\usepackage{IEEEtrantools,stackengine}
\stackMath

\usepackage{authblk}
\usepackage{fullpage}
\usepackage{amssymb,amsmath}
\usepackage{adjustbox}
\usepackage{graphicx}
\usepackage{longtable}
\usepackage{subcaption}
\usepackage{tcolorbox}
\tcbuselibrary{breakable}
\usepackage{float}

\usepackage{graphicx}
\usepackage{subcaption}
\usepackage{tikz}
\usetikzlibrary{shapes.callouts,positioning,calc}

\usepackage{booktabs}
\usepackage{pdflscape}
\usepackage{array}

\usepackage{siunitx}
\usepackage{csquotes}

\usepackage{graphicx}
\usepackage{tikz}
\usetikzlibrary{calc}

\usepackage{xspace}

\usepackage{booktabs}
\usepackage{longtable}
\usepackage{multirow}

\usepackage[dvipsnames, table]{xcolor}

\definecolor{darkgreen}{rgb}{0.0, 0.5, 0.0}

\definecolor{lightyellow}{HTML}{FFE699}
\definecolor{red_revision}{HTML}{FF0000}

\usepackage{setspace}
\usepackage[unicode=true]{hyperref}
\hypersetup{breaklinks=true,
            pdfborder={0 0 0},
            allcolors=black}
\usepackage[%
  sort&compress
]{cleveref}
  \Crefname{appendix}{Supplement}{Supplements}
  \Crefname{figure}{Fig.}{Fig.}

\usepackage{times}
\usepackage{enumerate}
\usepackage{enumitem}
\usepackage{float}

\usepackage{rotating}
\usepackage{tabularx}
\usepackage{dcolumn}
\usepackage{pdflscape}
\usepackage{rotating}

\usepackage{array}

\newcommand\autoagent{\textsc{AutoSynthesis}\xspace}

\usepackage{titlesec}

\titleformat{\subsubsection}
  {\normalfont\itshape}{\thesubsubsection}{1em}{}

\titleformat{\subsection}
  {\normalfont\bfseries}{\thesubsection}{1em}{}

\makeatletter
\renewcommand{\fps@figure}{H}         
\renewcommand{\fps@table}{H}         
\makeatother 

\renewcommand{\arraystretch}{1.2}

\allowdisplaybreaks

\usepackage{eurosym}
\usepackage{comment}

\usepackage{ntheorem}
\theoremseparator{:}

\usepackage{makecell}

\begin{document}


\title{\centering\LARGE\singlespacing \autoagent: An agentic system for automated meta-analysis}

\renewcommand\Affilfont{\fontsize{9}{10.8}\selectfont}

\author[1]{Moein Taherinezhad}
\author[2,3]{Sebastian Maier}
\author[4]{Gerardo Vitagliano}
\author[1]{Francesco Pierri\textsuperscript{$\dagger$}}
\author[2,3]{Stefan Feuerriegel\textsuperscript{$\dagger$}}

\begingroup
\renewcommand{\thefootnote}{\fnsymbol{footnote}}
\footnotetext[2]{Joint supervision and corresponding authors: \href{mailto:francesco.pierri@polimi.it}{francesco.pierri@polimi.it}, \href{mailto:feuerriegel@lmu.de}{feuerriegel@lmu.de} }
\endgroup

\affil[1]{Politecnico di Milano}
\affil[2]{LMU Munich, Munich, Germany}
\affil[3]{Munich Center for Machine Learning, Munich, Germany}
\affil[4]{MIT CSAIL}

\date{}

\maketitle

\newpage

\begin{abstract}\normalfont
\noindent
Evidence synthesis is crucial for turning primary research into reliable knowledge for science, medicine, education, and policy. Yet, quantitative evidence synthesis remains largely manual and difficult to scale. Here, we introduce \autoagent, an end-to-end multi-agent system for automated meta-analysis. Given a research question in natural language, \autoagent formulates a search strategy, retrieves scientific literature, screens candidate studies, assesses full-text eligibility, extracts quantitative statistics, computes standardized effect sizes, and finally performs random-effects meta-analysis. \autoagent further supports heterogeneity analysis to examine how effect sizes vary across moderators, as well as risk-of-bias assessment. As output, \autoagent produces a transparent report aligned with PRISMA guidelines. 
In our application, \autoagent screened over 28 studies and extracted more than 20 quantitative claims. The pooled effect estimates produced by \autoagent are similar to 
Hedges’ $g$ of expert-conducted meta-analyses, indicating close agreement with manual evidence synthesis. Together, these results show that \autoagent can make quantitative evidence synthesis more scalable, thereby supporting evidence-based decision-making across disciplines.
\end{abstract}

\flushbottom
\maketitle
\thispagestyle{empty}


\sloppy
\raggedbottom


\newpage

\section{Introduction}
\label{sec:introduction}


Systematic meta-analyses play a central role in evidence-based research by synthesizing findings from multiple independent studies into quantitative conclusions \cite{Gurevitch2018,Page2021}. By integrating evidence across diverse studies, they increase statistical power, improve the precision of effect estimates, and enable researchers to identify sources of heterogeneity that may not be apparent in individual studies. As a result, meta-analyses are used widely across disciplines, including clinical medicine (e.g., for developing evidence-based treatment guidelines \cite{Higgins2024}), psychology (e.g. for evaluating behavioral and mental health interventions \cite{nosek2015,Flake2022,Chen2025}), education (e.g., for assessing the effectiveness of learning interventions \cite{Holzner2025,Arifin2025,Tlili2025}), public health (e.g., for informing health policy and preventive interventions \cite{Bayat2025,OMahoney2025}), and computer science (e.g., for synthesizing the effect of human--computer interaction technologies on user behavior \cite{Hlbling2025}). Meta-analyses therefore play an important role in building the evidence base to inform policy decisions.


However, producing a high-quality meta-analysis is a manual process that is expensive and time-consuming. A typical meta-analysis requires extensive effort to formulate search strategies, retrieve literature from multiple databases, screen hundreds or thousands of candidate studies, assess eligibility, extract quantitative data, compute standardized effect sizes, perform statistical synthesis, and prepare reporting according to guidelines such as PRISMA \cite{Page2021}. Previous studies estimate that the systematic review alone often requires more than one thousand person-hours and may take more than a year to complete \cite{Borah2017,Bastian2010}, creating a substantial lag between the publication of primary studies and the availability of synthesized evidence. Here, we present an agentic framework for automatic meta-analysis to synthesize evidence in a reproducible, scalable, near real-time manner. 


Recent advances in large language models (LLMs) have substantially expanded the scope of AI-assisted scientific research. Owing to their ability to reason over unstructured scientific documents, generate executable code, and synthesize information across diverse sources \cite{Feuerriegel2023}, LLMs have recently been applied to a wide range of scientific workflows, including hypothesis generation, scientific discovery, software engineering, automated reproducibility assessments, and data analysis \cite{Gottweis2026,Holtdirk2026,yamada2025}. Existing research on LLM-assisted evidence synthesis broadly falls into two categories. The first uses LLMs for literature retrieval, screening, and eligibility assessment, in order to identify studies and research questions relevant to a user query. \cite{Syriani2024,Guo2024,Boateng2024}. The second uses LLMs for structured information extraction, such as extracting the outcomes or other measures \cite{Forster2026,ahad2024,Li2026, Rouzrokh2025,xu2025,Padarha2026}. Some of these systems can even perform systematic reviews \cite{Chen2026}. In contrast, end-to-end systems that move beyond screening and extraction to compute standardized effect sizes and perform complete statistical meta-analysis remain missing.

\begin{figure}[H]
    \centering
    \includegraphics[width=\linewidth]{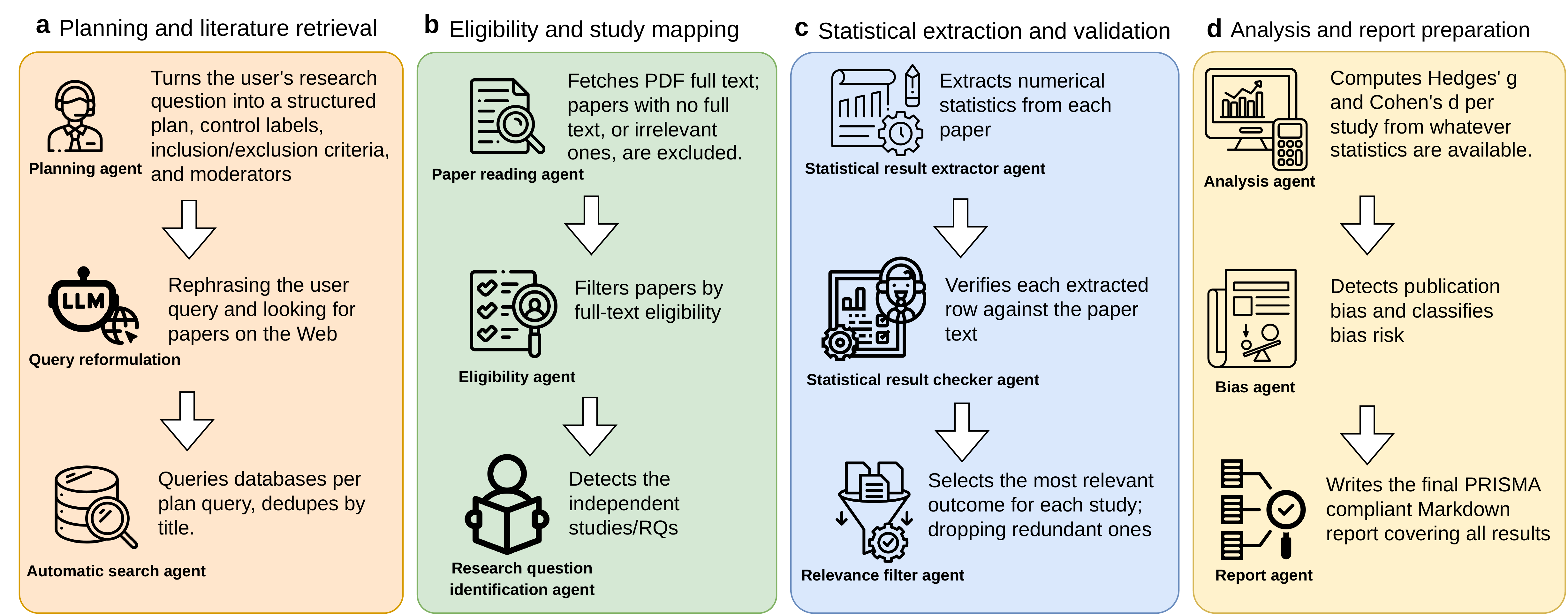}
    \caption{\textbf{Overview of the \autoagent framework.} A multi-agent framework including: \textbf{a}, The workflow begins by transforming the user's research question into a structured review protocol and retrieving candidate studies from multiple scientific databases using LLM-generated search queries. \textbf{b}, Retrieved articles undergo full-text retrieval, eligibility assessment, and study mapping, where independent studies and research questions are identified within each paper. \textbf{c}, Statistical evidence is extracted, validated against the source article, and filtered to retain only the results relevant to the target meta-analysis before being standardized into a common representation. \textbf{d}, Standardized effect sizes are synthesized using a random-effects meta-analysis, followed by publication-bias assessment and automatic generation of a PRISMA-compliant report together with visualizations and complete audit records. The modular architecture enables every stage of the evidence-synthesis process to be independently inspected, verified, and compared with expert-conducted meta-analysis.}
    \label{fig:overview}
\end{figure}

In this paper, we present \autoagent, an end-to-end multi-agent system for automated meta-analysis (Fig.~\ref{fig:overview}). \autoagent takes a research question in natural language as input. Then, \autoagent autonomously formulates a search strategy, retrieves the literature from dedicated scientific databases, screens the retrieved papers, assesses the eligibility based on the full text, extracts quantitative data (in addition to study metadata), computes standardized effect sizes (such as Cohen's $d$ and Hedges' $ g$), and performs a random-effects meta-analysis. As part of this workflow, \autoagent estimates and reports between-study heterogeneity using standard statistics such as $I^2$ and $\tau^2$. When specified by the user, \autoagent can also perform heterogeneity analysis to understand how the effect sizes vary across different moderators of interest. In addition, \autoagent conducts a principled reporting bias assessment according to the best practice (e.g., trim-and-fill \cite{Duval2000} procedure, checks for small-study effects such as through funnel plot analyses \cite{Egger1997} to test for potential publication bias), as well as a risk of bias in included studies analysis (e.g., RoB~2 for randomized trials \cite{Sterne2019} and ROBINS-I for non-randomized studies of interventions \cite{Sterne2016}). The output is a complete report following PRISMA guidelines \cite{Page2021}. For this, \autoagent combines several specialized LLM agents within a rigorous statistical workflow to support complex reasoning over scientific evidence. \autoagent records the complete agent traces and intermediate decisions during the workflow, which enables end-users to inspect and audit each step of the evidence synthesis process (e.g., by inspecting the reasons for inclusion/exclusion of a specific study). In sum, \autoagent provides a transparent, reproducible, and scalable approach for evidence synthesis.

We demonstrate the capabilities of \autoagent by performing several automated meta-analyses across different scientific domains, including human--AI interaction, education, and psychology. To benchmark automated evidence synthesis, we compare the results produced by \autoagent against established expert-conducted meta-analyses used as reference cases \cite{Hlbling2025,Holzner2025}. Across these applications, \autoagent retrieved 28 candidate records, screened 19 papers for inclusion, and extracted quantitative evidence from 8 eligible studies. We empirically assess performance across multiple stages of the evidence synthesis process, including the overlap with expert-conducted reviews in literature retrieval and research questions selection, the accuracy of quantitative data extraction, and the agreement of the final effect sizes. Across all reference cases, the pooled effect estimates produced by \autoagent were within {$\pm 0.12$} Hedges' $g$ of the corresponding expert-conducted meta-analyses, indicating strong agreement with manual evidence synthesis across domains.

More broadly, automated meta-analysis through \autoagent can expand how quantitative evidence synthesis is produced and used. First, \autoagent can make meta-analysis cheaper and more scalable, lowering the barrier to quantitative evidence synthesis in settings where manual evidence synthesis would be too costly or time-consuming, thereby promoting evidence-based decision-making at scale in placed that would otherwise be guided by ad hoc judgment. Second, \autoagent may enable near-real-time synthesis of evidence. In this way, it extends the idea of ``living'' systematic reviews \cite{Iversen2026} to ``living'' meta-analyses, which can continuously integrate new evidence as studies appear. Third, automated workflows can make it easier to consistently apply established best practices  (e.g., such as risk-of-bias assessment \cite{Higgins2024} and PRISMA-aligned reporting \cite{Page2021}). Hence, \autoagent may help improve methodological rigor, especially for researchers without specialized expertise in meta-analysis. Together, these capabilities can support evidence-based decision-making in areas where timely and reliable summaries of the literature are needed, including clinical practice, education, and public policy \cite{Tyler2023}.


\section{Results}
\label{sec:results}

\subsection{\autoagent can perform an end-to-end meta-analysis}
\label{sec:illustrative_results}


To demonstrate the capabilities of \autoagent, we tasked the system with a meta-analytic research question on whether LLMs are more persuasive than humans. Evidence for this research question is contested \cite{Salvi2025, Bai2025, Matz2024, Hackenburg2024}, but increasingly relevant in light of ongoing debates about the persuasive power of generative AI with implications for psychology, communication, and society. A recent meta-analysis by Hölbling et al. (2025) \cite{Hlbling2025} provides a useful reference point ($k = 11$ effect sizes, 17,422 participants, which allows us to later compare the automated synthesis from \autoagent against a human benchmark.

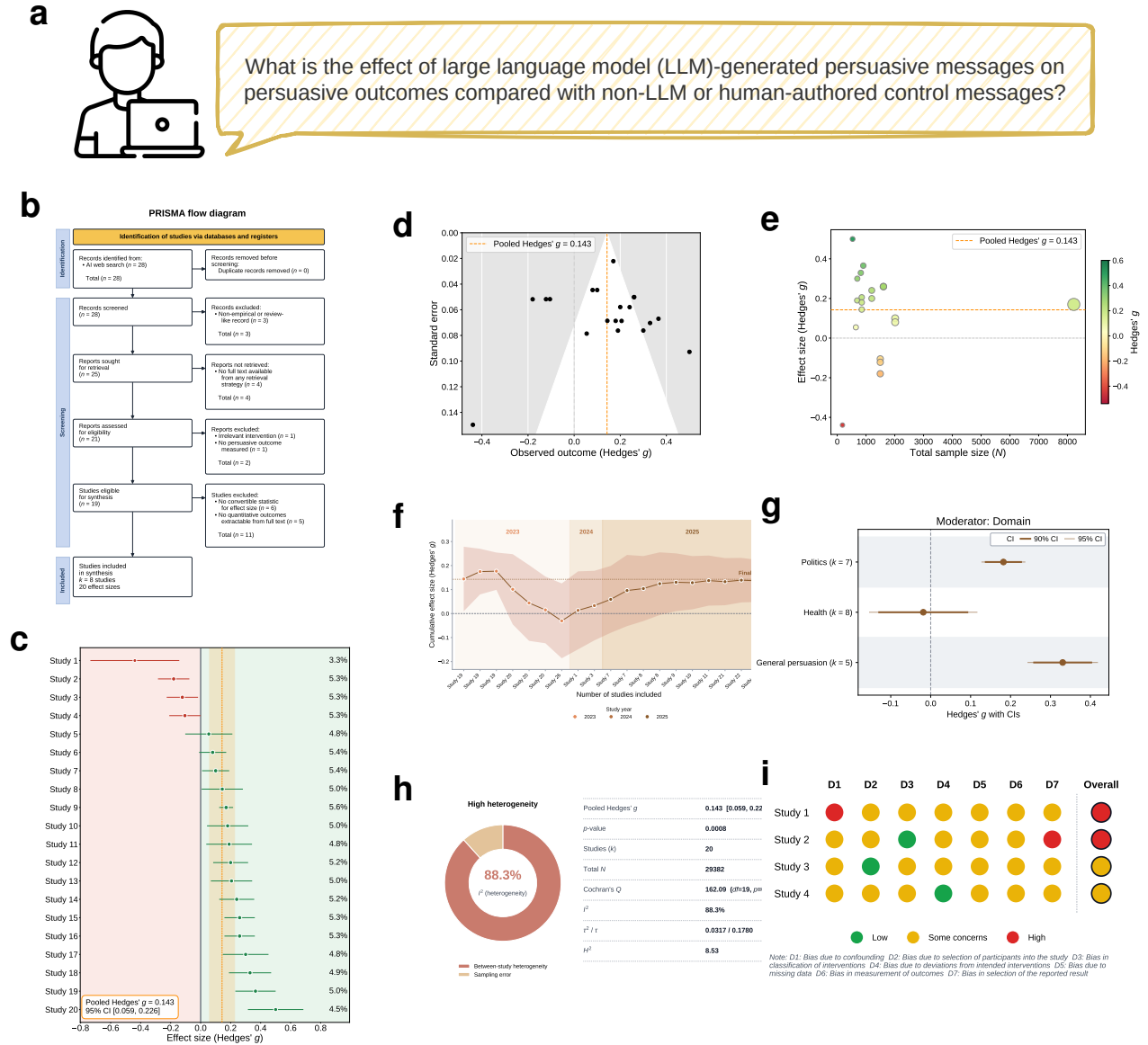
\begin{figure}[H]
    \centering

    \begin{tikzpicture}
        \node[inner sep=0, outer sep=0] (dash)
            {\includegraphics[width=0.92\textwidth]{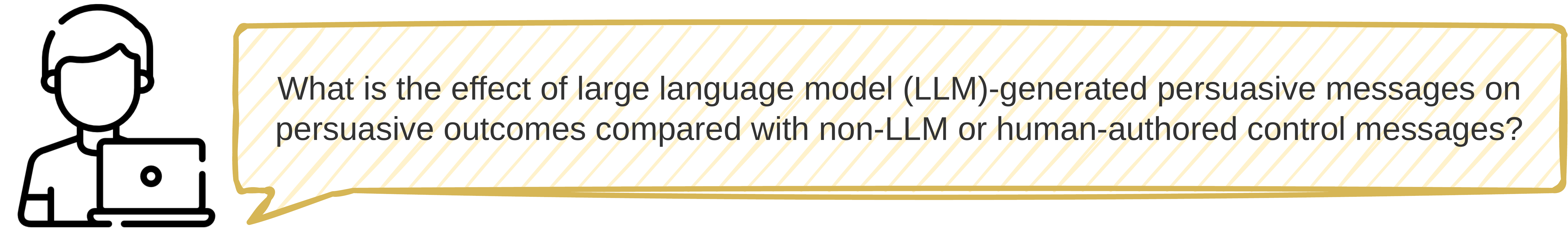}};
        \node[anchor=north west, font=\sffamily\bfseries\large, xshift=-18pt, yshift=5pt]
            at (dash.north west) {a};
    \end{tikzpicture}

    \vspace{0.6em}

    \noindent
    \begin{minipage}[t]{0.32\textwidth}
        \centering

        \begin{tikzpicture}
            \node[inner sep=0, outer sep=0] (bimg)
                {\includegraphics[width=\linewidth,height=0.27\textheight,keepaspectratio]{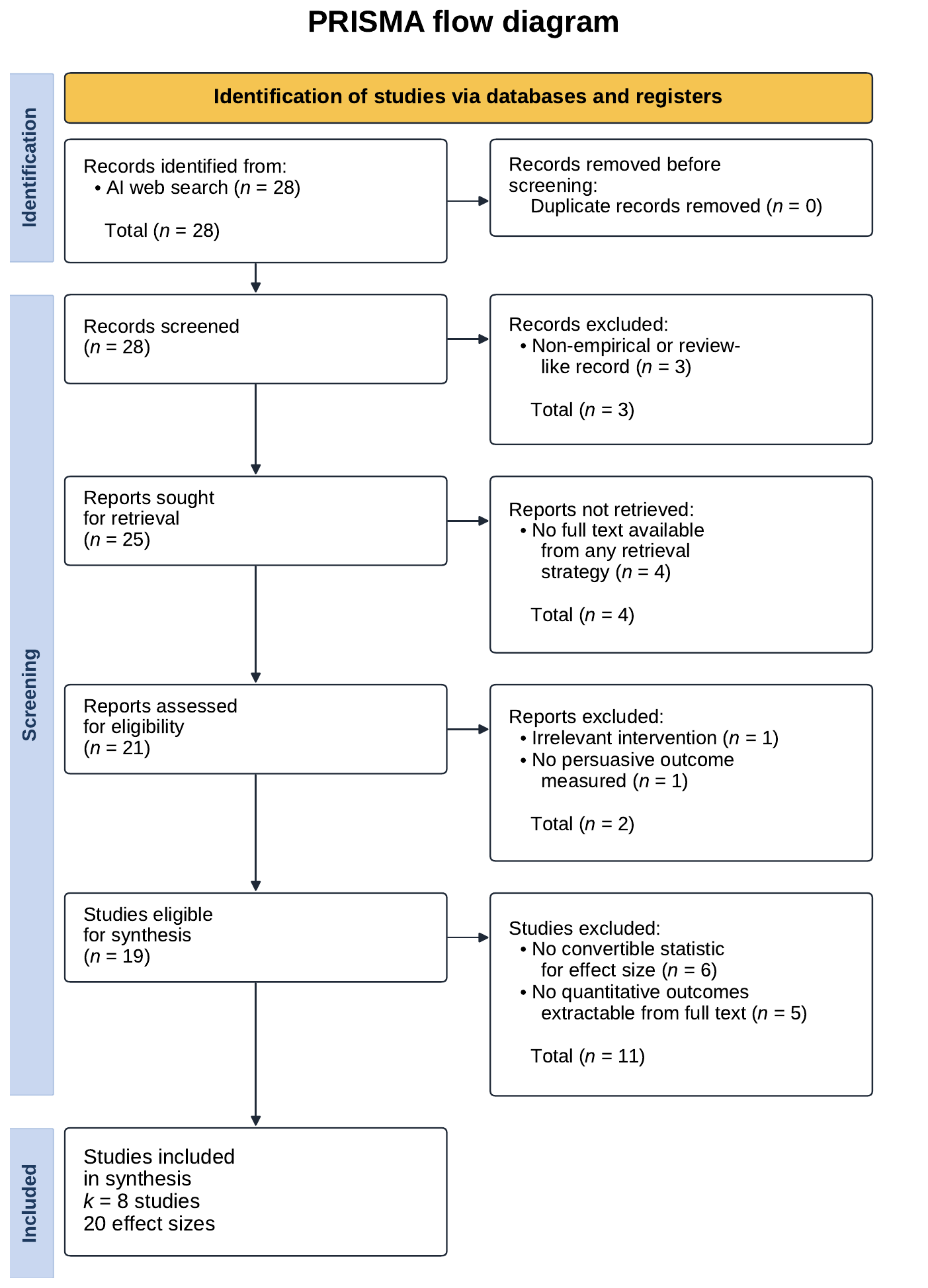}};
            \node[anchor=north west, font=\sffamily\bfseries\large, xshift=-18pt, yshift=10pt]
                at (bimg.north west) {b};
        \end{tikzpicture}

        \vspace{0.45em}

        \begin{tikzpicture}
            \node[inner sep=0, outer sep=0] (cimg)
                {\includegraphics[width=\linewidth,height=0.27\textheight,keepaspectratio]{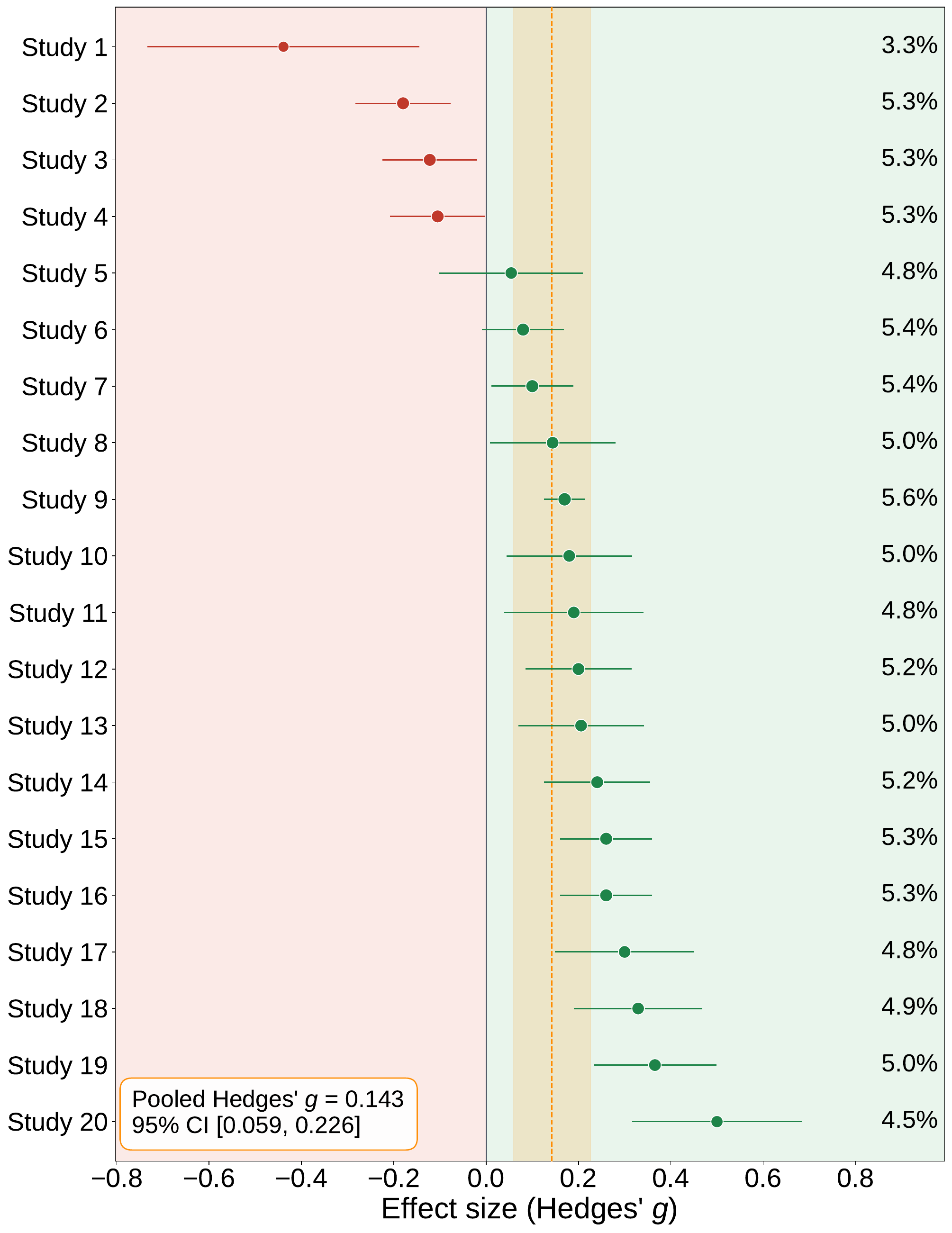}};
            \node[anchor=north west, font=\sffamily\bfseries\large, xshift=-18pt, yshift=10pt]
                at (cimg.north west) {c};
        \end{tikzpicture}
    \end{minipage}
    \hfill
    \begin{minipage}[t]{0.66\textwidth}
        \centering
        \setlength{\tabcolsep}{1pt}

        \begin{tabular}{@{}p{0.49\linewidth}p{0.49\linewidth}@{}}
            \begin{tikzpicture}
                \node[inner sep=0, outer sep=0] (dimg)
                    {\includegraphics[width=\linewidth,height=0.16\textheight,keepaspectratio]{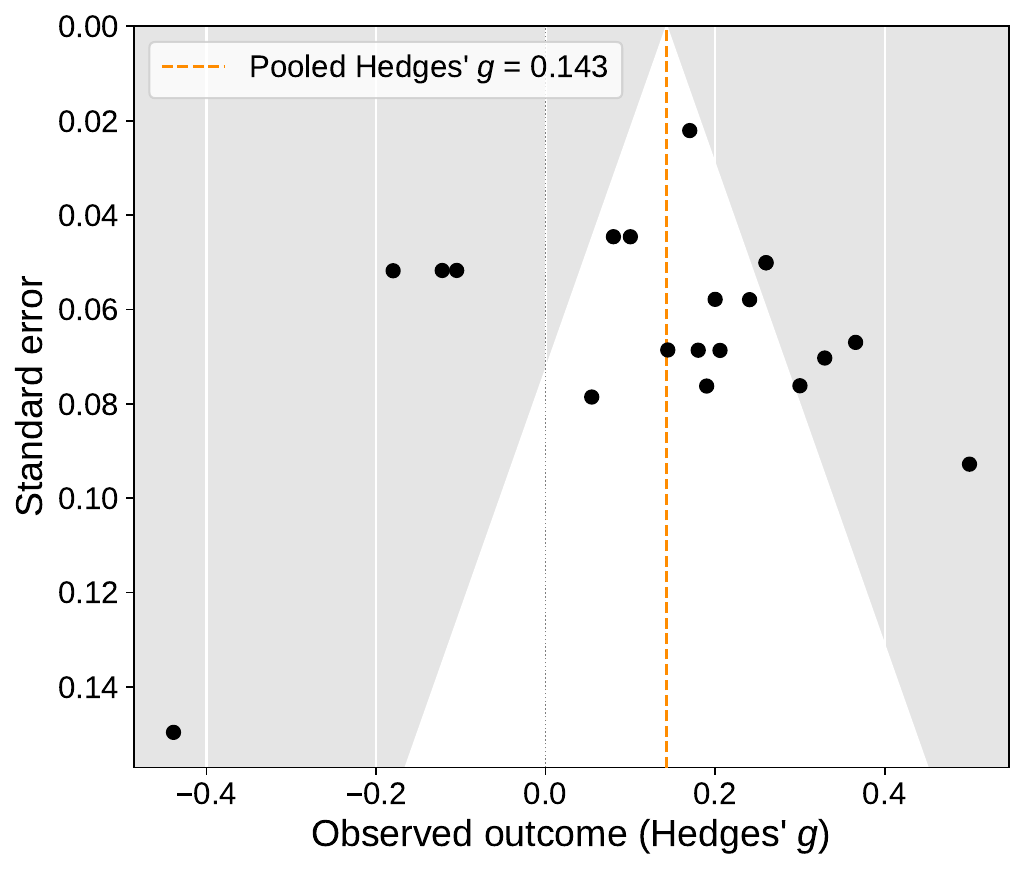}};
                \node[anchor=north west, font=\sffamily\bfseries\large, xshift=-18pt, yshift=10pt]
                    at (dimg.north west) {d};
            \end{tikzpicture}
            &
            \begin{tikzpicture}
                \node[inner sep=0, outer sep=0] (eimg)
                    {\includegraphics[width=\linewidth,height=0.16\textheight,keepaspectratio]{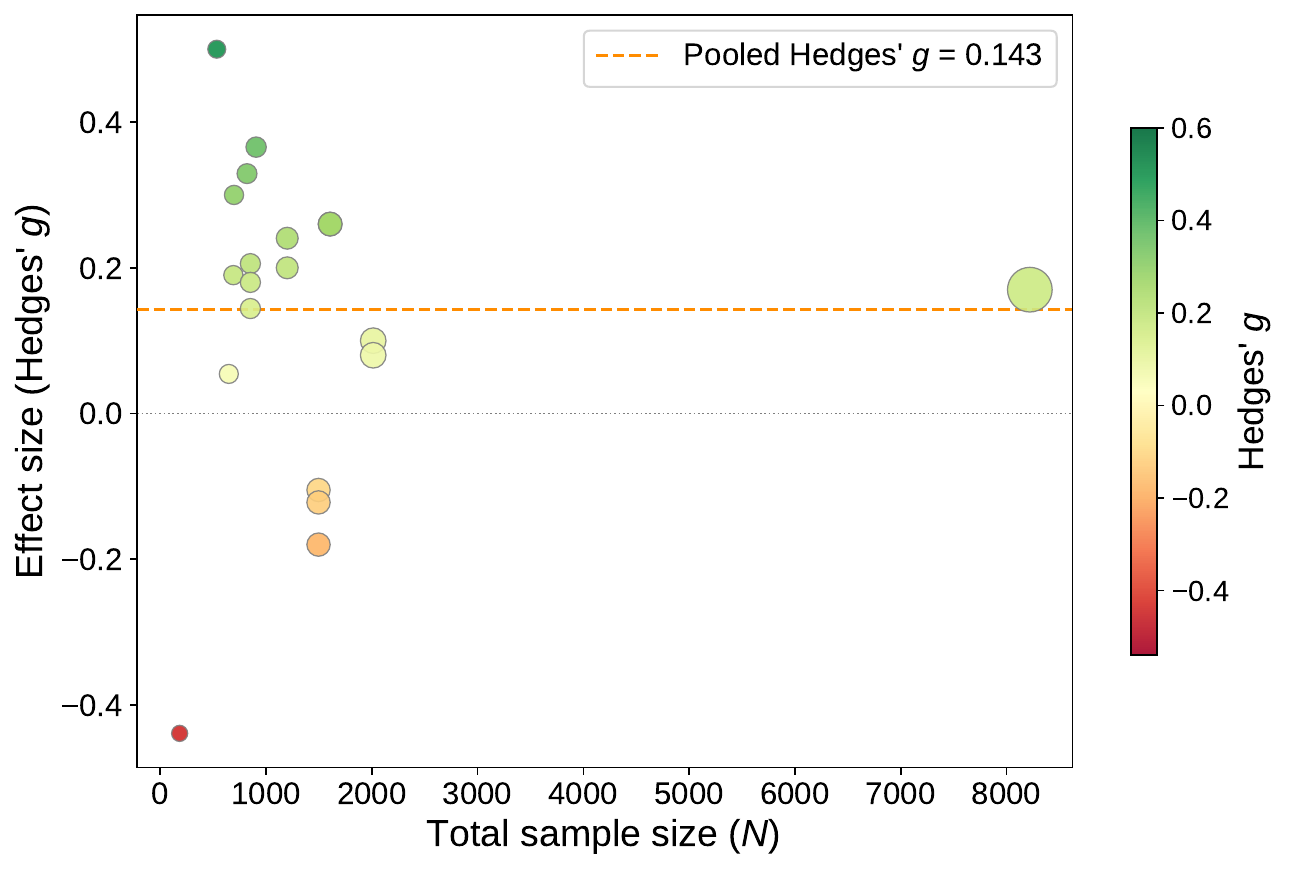}};
                \node[anchor=north west, font=\sffamily\bfseries\large, xshift=-18pt, yshift=10pt]
                    at (eimg.north west) {e};
            \end{tikzpicture}
            \\[0.45em]

            \begin{tikzpicture}
                \node[inner sep=0, outer sep=0] (fimg)
                    {\includegraphics[width=\linewidth,height=0.16\textheight,keepaspectratio]{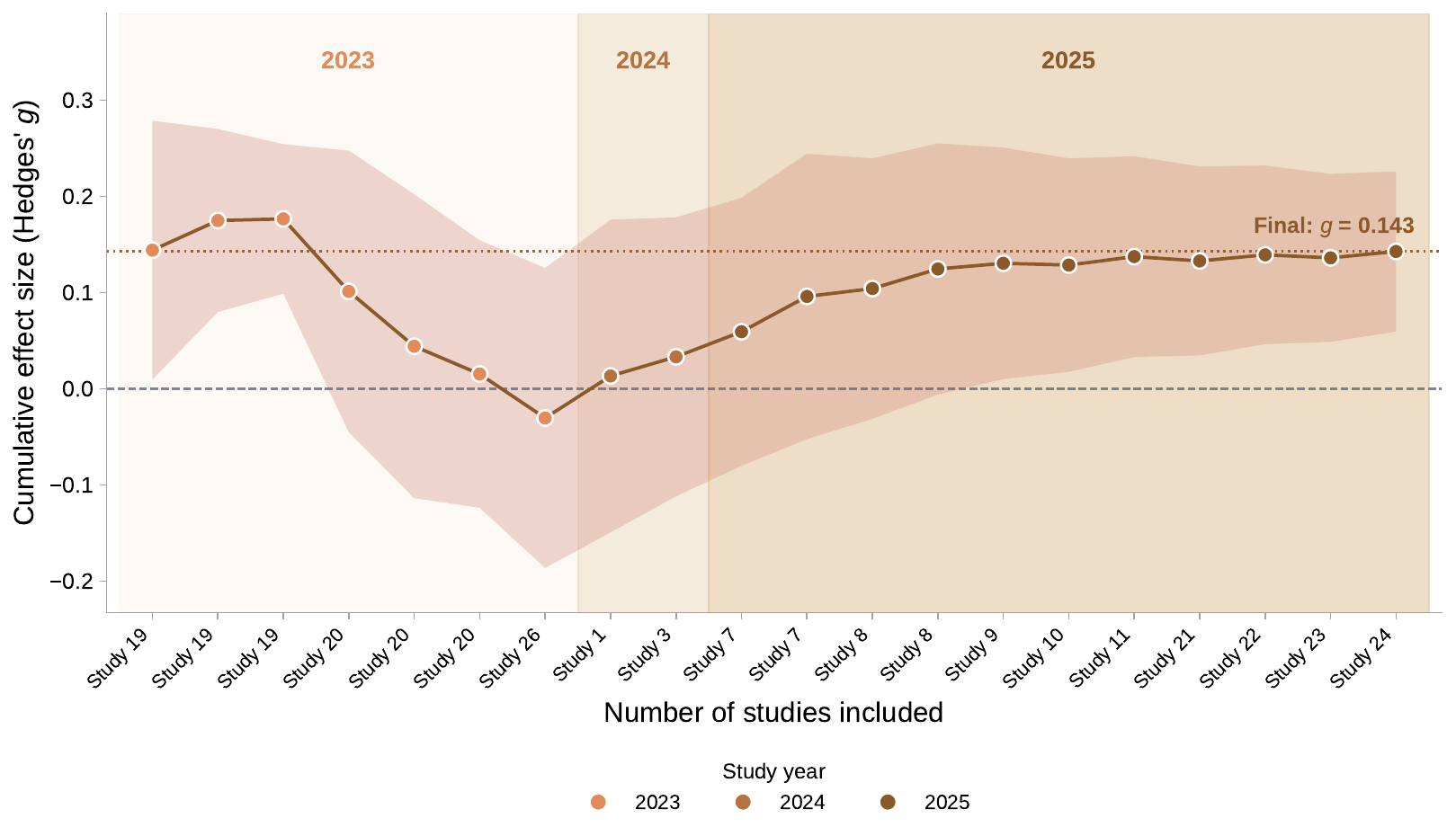}};
                \node[anchor=north west, font=\sffamily\bfseries\large, xshift=-18pt, yshift=10pt]
                    at (fimg.north west) {f};
            \end{tikzpicture}
            &
            \begin{tikzpicture}
                \node[inner sep=0, outer sep=0] (gimg)
                    {\includegraphics[width=\linewidth,height=0.16\textheight,keepaspectratio]{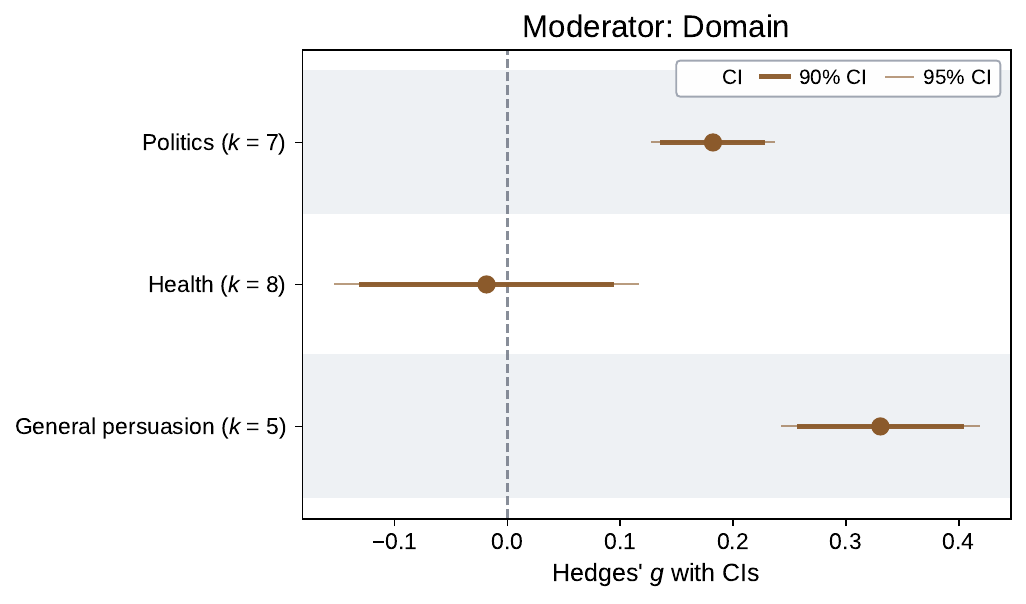}};
                \node[anchor=north west, font=\sffamily\bfseries\large, xshift=0pt, yshift=10pt]
                    at (gimg.north west) {g};
            \end{tikzpicture}
            \\[0.45em]

            \begin{tikzpicture}
                \node[inner sep=0, outer sep=0] (himg)
                    {\includegraphics[width=\linewidth,height=0.16\textheight,keepaspectratio]{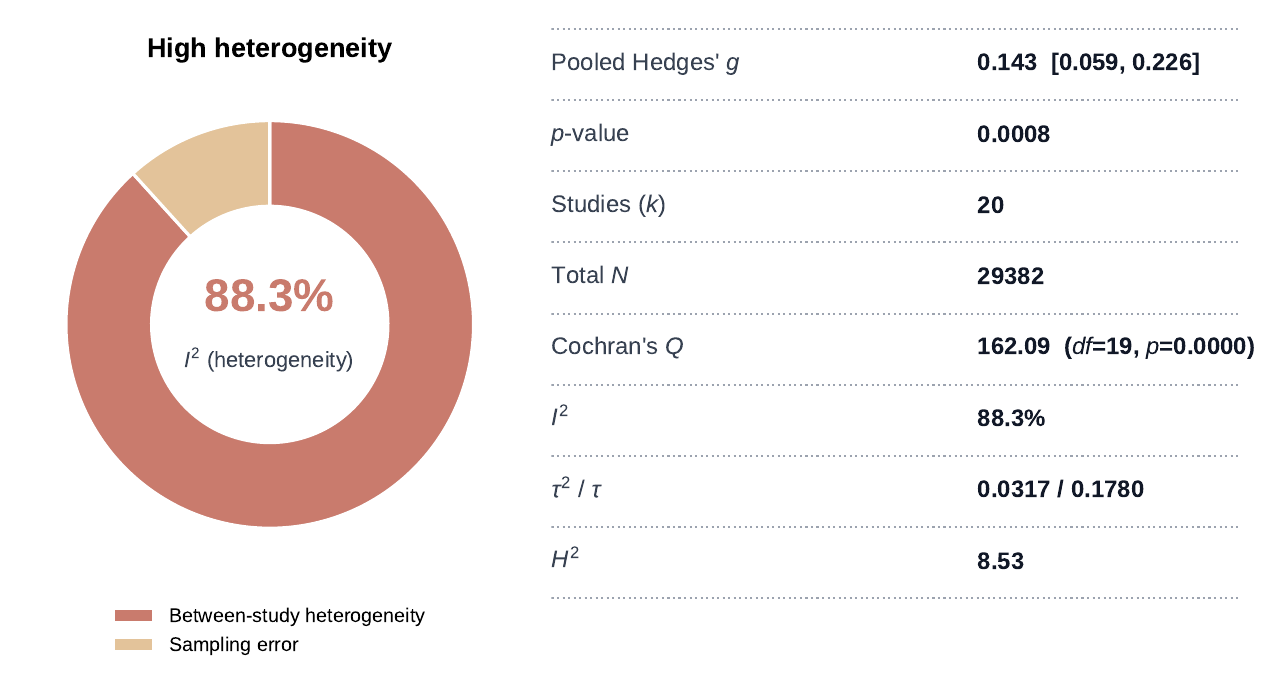}};
                \node[anchor=north west, font=\sffamily\bfseries\large, xshift=-18pt, yshift=10pt]
                    at (himg.north west) {h};
            \end{tikzpicture}
            &
            \begin{tikzpicture}
                \node[inner sep=0, outer sep=0] (iimg)
                    {\includegraphics[width=\linewidth,height=0.16\textheight,keepaspectratio]{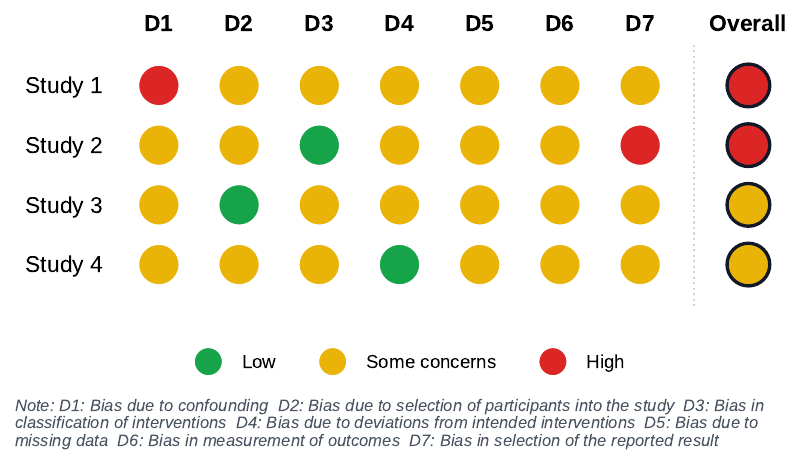}};
                \node[anchor=north west, font=\sffamily\bfseries\large, xshift=-5pt, yshift=10pt]
                    at (iimg.north west) {i};
            \end{tikzpicture}
        \end{tabular}
    \end{minipage}

    \caption{\footnotesize \textbf{Demonstration of automated evidence synthesis using \autoagent.}
    To demonstrate the capabilities of \autoagent, we tasked the framework with synthesizing evidence on the persuasive power of large language models, for which evidence is contested following the research question of the recent meta-analysis by Hölbling et al. (2025)~\cite{Hlbling2025}.
    \textbf{a,}~The input to \autoagent is a user-provided research question to initiate the automated evidence synthesis workflow. As a result, \autoagent produces the following output:
    \textbf{b,}~PRISMA 2020 \cite{Page2021} flow diagram summarizing the screening process and the final study inclusion.
    \textbf{c,}~The forest plot shows the study-level effects (in terms of Hedges' $g$) and the pooled random-effects estimate with 95\% confidence intervals (CIs).
    \textbf{d,}~Funnel plot to assess small-study effects and potential publication bias.
    \textbf{e,}~Scatter plot as a diagnostic tool to inspect the relationship between effect sizes and sample sizes.
    \textbf{f,}~The cumulative plot shows how the pooled random-effects estimate evolves as studies accumulate.
    \textbf{g,}~The moderator analysis shows the pooled effect sizes across predefined study-level characteristics (here: the domain of the underlying persuasion study).
    \textbf{h,}~Heterogeneity statistics summarizing the between-study variability (including $I^2$, $\tau^2$, and Cochran's $Q$).
    \textbf{i,}~Study-level risk-of-bias assessment following ROINS-I \cite{Sterne2016}, where the assessment is varies from low to high concern.
    \autoagent further produces a study-level overview with metadata and extracted effect sizes in a tabular format; see (Table. ~\ref{tab:study_table_export},~\ref{tab:outcome_table_export})}
    \label{fig:dashboard}
    
\end{figure}

The input to \autoagent was the following research question (see Fig.~\ref{fig:dashboard}a): ``\emph{What is the effect of large language model (LLM)-generated persuasive messages on persuasive outcomes compared with non-LLM or human-authored control messages?}'' We did not specify a start or end date for exclusion. Using this input, \autoagent performed a complete evidence synthesis: \autoagent first produced a structured review protocol, retrieved and screened eligible studies, extracted statistical results, computed standardized effect sizes, and performed a random-effects meta-analysis. The final output is a structured report, which includes a PRISMA 2020 flow diagram, a forest plot, the pooled meta-analytic effect estimate in terms of Hedges' $g$ with accompanying statistics (e.g., such as $I^2$, $\tau^2$, and Cochran’s $Q$), a funnel plot for publication bias diagnostics, and risk-of-bias assessments (Fig.~\ref{fig:dashboard}b--i). In addition, \autoagent produces a narrative Markdown report and a study-level overview table summarizing the included studies, relevant metadata, and the extracted effect sizes (see Supplementary~Table~\ref{tab:study_table_export}).

For this research question, \autoagent retrieved 28 candidate records, retained 25 after title and abstract screening, identified 19 eligible studies based on the full text, and ultimately included 8 studies with overall $k=20$ effect size estimates (Fig.~\ref{fig:dashboard}b). The resulting random-effects meta-analysis estimated a pooled effect of Hedges' $g = 0.143$ (95\% CI $[0.059, 0.226]$, $p < 0.001$; Fig.~\ref{fig:dashboard}c), which suggests that LLM-generated persuasive messages were, on average, more persuasive than the control conditions. At the same time, the analysis indicated substantial between-study heterogeneity ($I^2 = 88.3\%$; Fig.~\ref{fig:dashboard}h). Additional diagnostics generated by \autoagent supported this interpretation: the funnel plot indicated potential small-study effects (Fig.~\ref{fig:dashboard}d). To better inspect small-study effects, \autoagent also produced plots comparing the effect size against sample size to show how study-level estimates relate to statistical power (Fig.~\ref{fig:dashboard}e). A leave-one-out sensitivity analysis further confirms that the overall pooled effect remained largely robust to the exclusion of an individual study (Fig.~\ref{fig:leave_one_out}). Further, \autoagent produced a cumulative meta-analysis showing how the pooled estimate evolved as evidence accumulated over time Fig.~\ref{fig:dashboard}f). This provides insights into how the persuasive power of LLMs increased as the LLM capabilities improved over different years. Moderator analysis further showed that effect sizes differed across application domains (Fig.~\ref{fig:dashboard}g). Finally, the risk-of-bias assessment highlighted that some studies may require cautious interpretation, for example, because of small samples or design limitations (Fig.~\ref{fig:dashboard}i).

\newpage
\thispagestyle{empty}

\newpage


\subsection{\autoagent can construct a relevant evidence base for meta-analyses}

To assess the literature retrieval process, we evaluated whether \autoagent constructed an appropriate evidence base for the meta-analysis. Here, our evaluation focuses on the intermediate steps of evidence synthesis, namely, study retrieval, screening, full-text eligibility assessment, and inclusion in the final quantitative synthesis.  

\begin{figure}[H]
    \centering

    \begin{minipage}[t]{0.40\linewidth}
        \centering
        \begin{tikzpicture}
            \node[inner sep=0] (img)
                {\includegraphics[width=\linewidth]{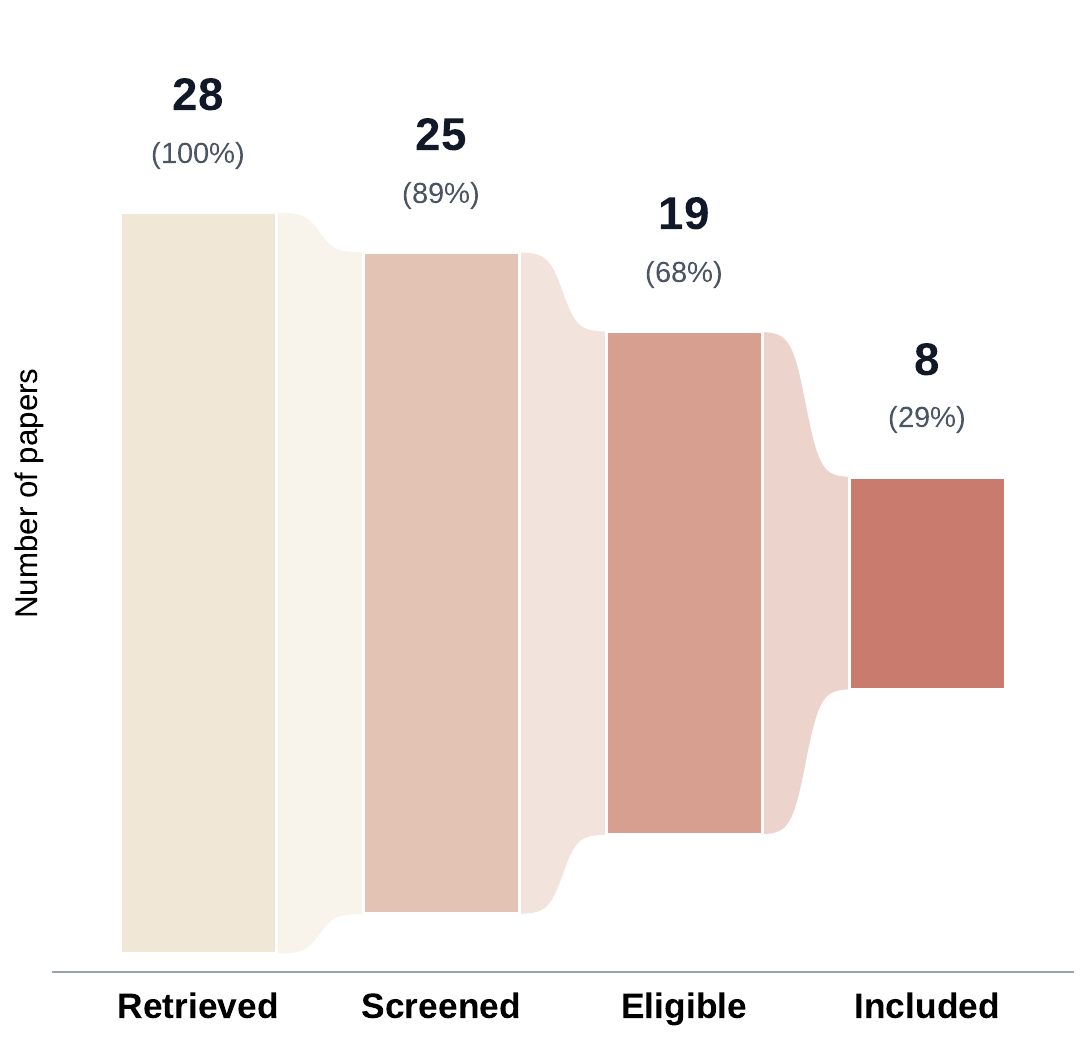}};
            \node[
                anchor=north west,
                font=\sffamily\bfseries\normalsize,
                xshift=4pt,
                yshift=50pt
            ] at (img.north west) {a};
        \end{tikzpicture}
    \end{minipage}
    \hfill
    \begin{minipage}[t]{0.49\linewidth}
        \centering
        \begin{tikzpicture}
            \node[inner sep=0] (img)
                {\includegraphics[width=\linewidth]{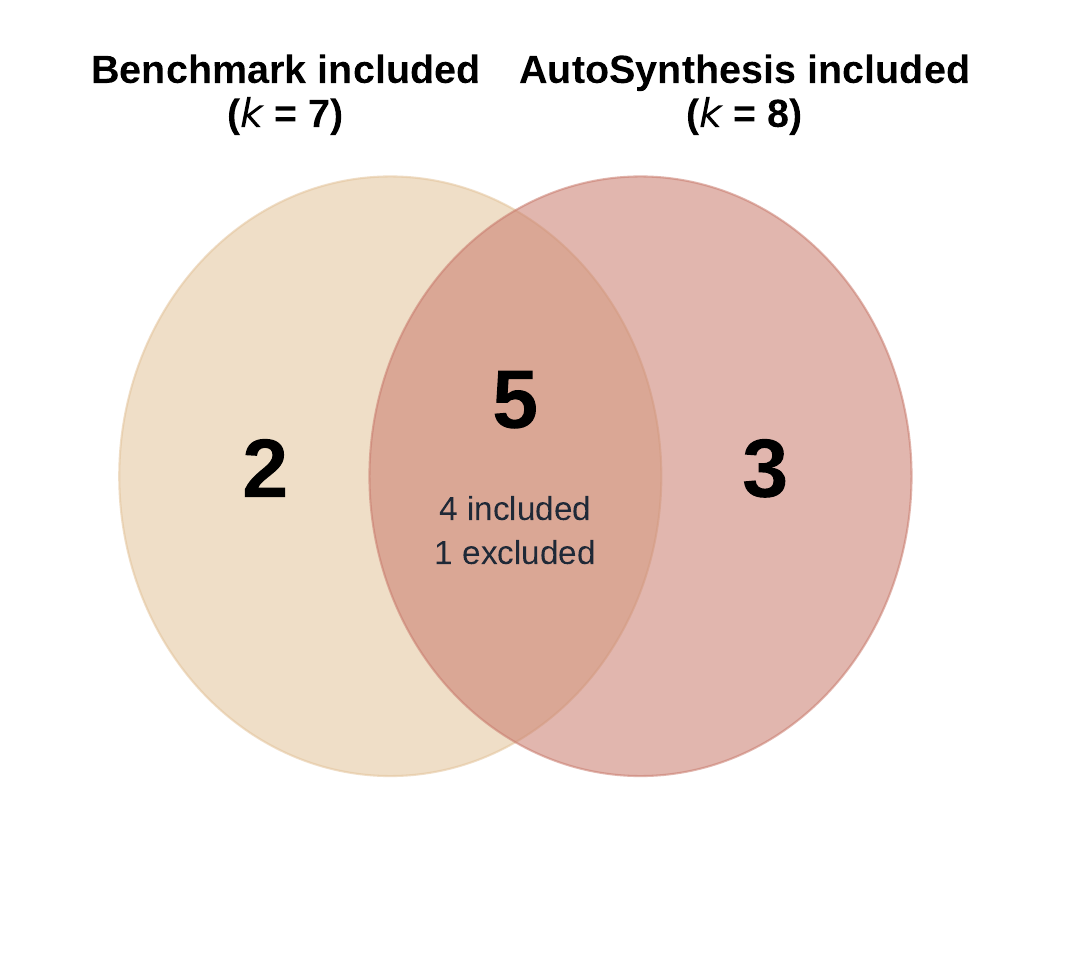}};
            \node[
                anchor=north west,
                font=\sffamily\bfseries\normalsize,
                xshift=4pt,
                yshift=30pt
            ] at (img.north west) {b};
        \end{tikzpicture}
    \end{minipage}

    \caption{\textbf{Evaluation of the literature retrieval and study selection.}
    \textbf{a,}~The funnel summarizing study selection process by reporting the number of studies retrieved, screened, eligible, and finally included.
    \textbf{b,}~Agreement between the studies included by \autoagent{} and those included
    in the benchmark meta-analysis. 
    }
    \label{fig:retrieval-eval}
\end{figure}

The study selection through \autoagent was as follows (Fig.~\ref{fig:retrieval-eval}a). \autoagent retrieved 28 candidate records, retained 25, after screening title and abstract, and identified 19 to satisfy the predefined eligibility criteria. Of these, 8 studies ultimately contributed to the quantitative synthesis after excluding papers that lacked sufficient quantitative information for effect size computation or statistical extraction. The resulting evidence base from \autoagent showed substantial agreement with the human benchmark (Fig.~\ref{fig:retrieval-eval}b); five of the seven studies included in the human meta-analysis were also included by \autoagent, corresponding to a recall of 71.4\%. At the same time, \autoagent included three additional studies that were not part of the benchmark evidence base, corresponding to a precision of 62.5\%.

Upon qualitative inspection, the differences between the benchmark evidence base and the evidence base constructed by \autoagent were mainly due to retrieval and eligibility decisions. Of the two studies included in the benchmark meta-analysis but not recovered by \autoagent, one was not identified during the search process, whereas the other could not be processed because its full-text PDF was not readily accessible for automated retrieval. Conversely, \autoagent included three studies that were not part of the benchmark. One of these studies was published after the search cut-off date of Hölbling et al. (2025) \cite{Hlbling2025}, and one was available as a ResearchGate preprint, which was not included among the search sources used in the benchmark review. The third study focused on preferences for LLM-generated messages rather than persuasive outcomes; a qualitative review by the lead author of the benchmark meta-analysis indicated that this study was not directly relevant to the target research question. We therefore interpret this case as an error in the eligibility assessment by \autoagent, likely because the article used an extensive framing around persuasion. After accounting for these qualitative differences, the corrected overlap with the benchmark was higher: \autoagent recovered six of the seven benchmark studies, corresponding to a corrected recall of 85.7\%, and seven of the eight studies included by \autoagent were judged relevant, corresponding to a corrected precision of 87.5\%.

\subsection{\autoagent recovers accurate meta-analytic effect size estimates}

We next evaluated whether \autoagent reproduces the quantitative conclusions of published manual meta-analyses. 

\begin{figure}[H]
    \centering

    \begin{subfigure}[t]{0.49\linewidth}
        \centering
        \begin{tikzpicture}
            \node[inner sep=0] (img)
                {\includegraphics[width=\linewidth]{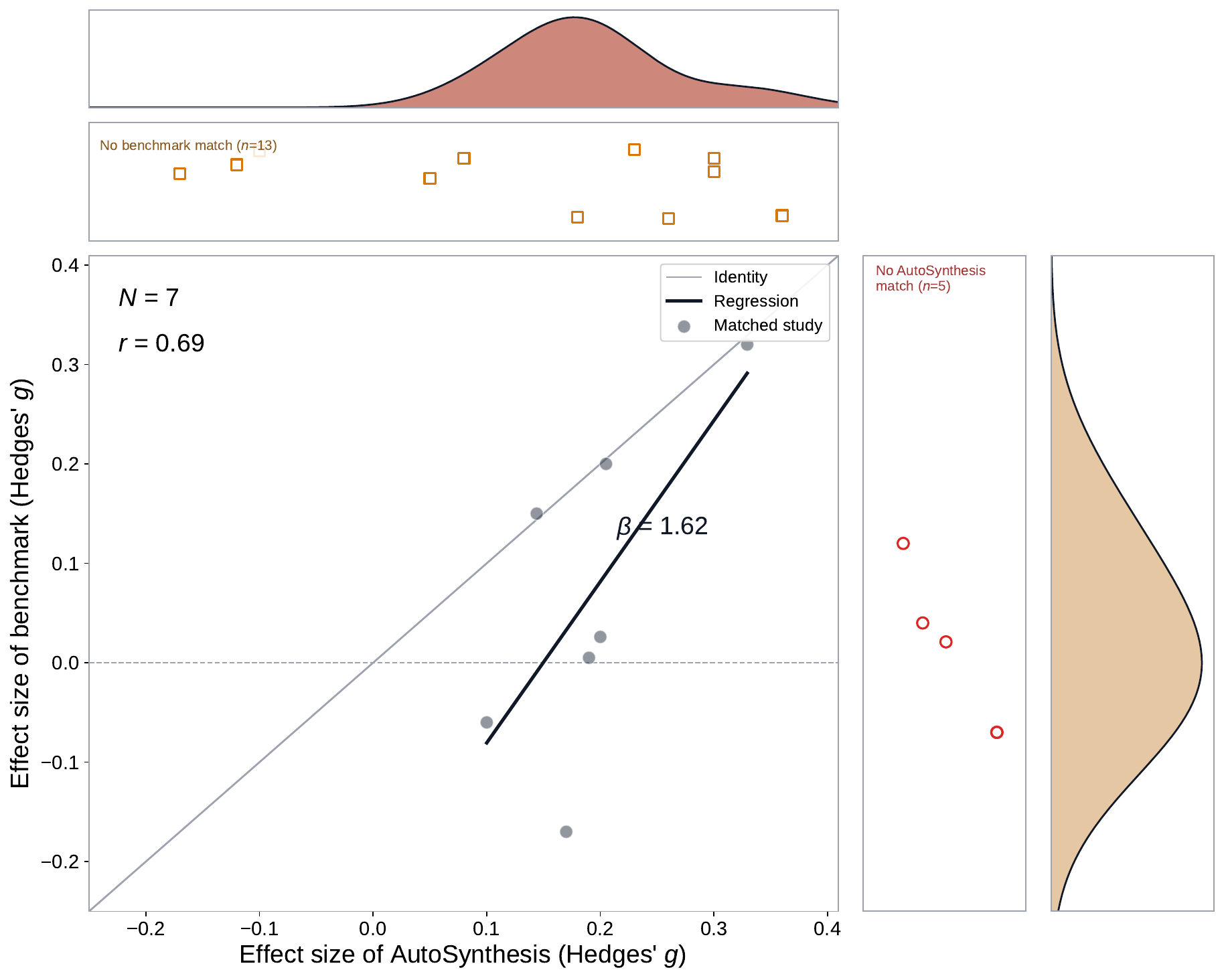}};
            \node[
                anchor=north west,
                font=\sffamily\bfseries\normalsize,
                xshift=4pt,
                yshift=45pt
            ] at (img.north west) {a};
        \end{tikzpicture}
    \end{subfigure}
    \hfill
    \begin{subfigure}[t]{0.49\linewidth}
        \centering
        \begin{tikzpicture}
            \node[inner sep=0] (img)
                {\includegraphics[width=\linewidth]{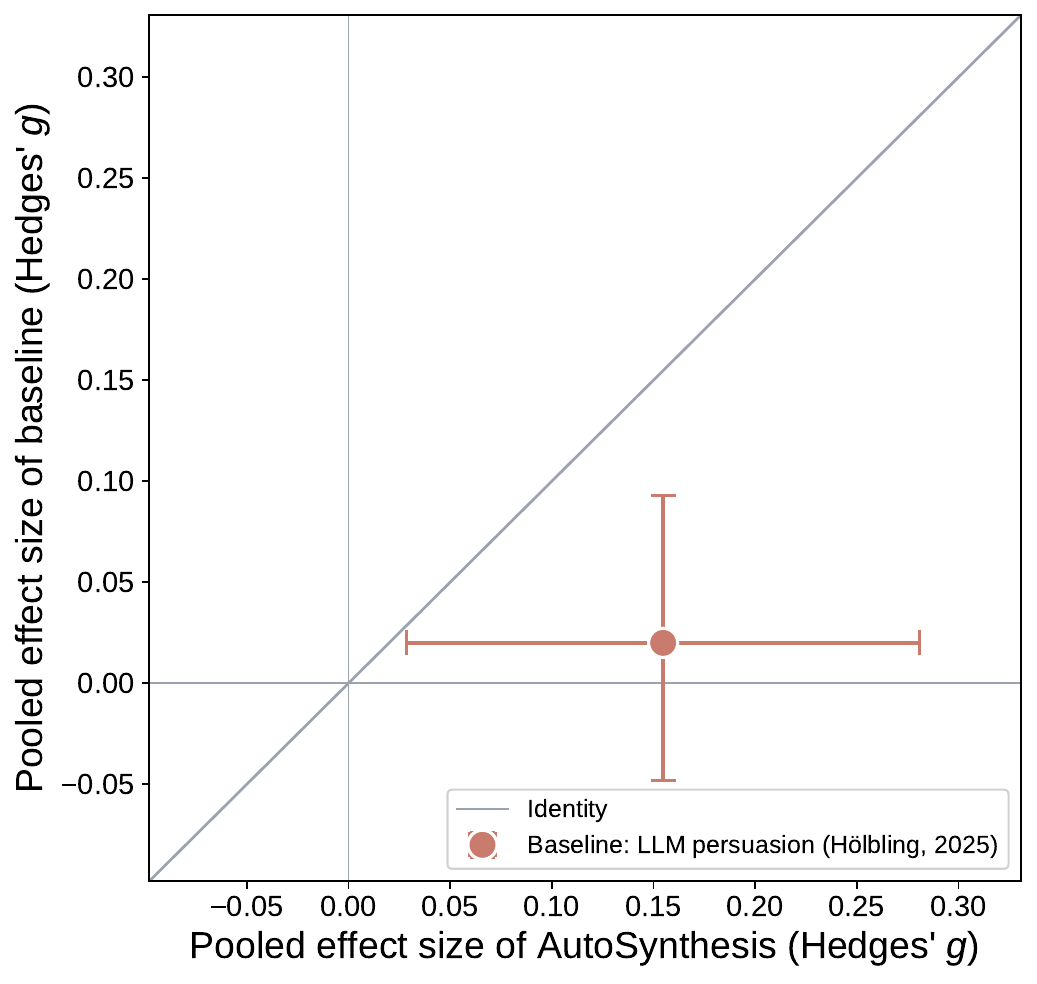}};
            \node[
                anchor=north west,
                font=\sffamily\bfseries\normalsize,
                xshift=4pt,
                yshift=10pt
            ] at (img.north west) {b};
        \end{tikzpicture}
    \end{subfigure}

    \caption{
    \textbf{Agreement between the meta-analytics estimates produced by \autoagent and the manual benchmark.}
    \textbf{a,}~Comparison of study-level standardized effect sizes (Hedges' $g$) extracted by \autoagent and those reported in the published human benchmark (gray points). The diagonal line indicates perfect agreement. Reported is the Pearson’s correlation coefficient ($r$) and the slope of the ordinary least squares (OLS) trend line ($\beta$). The separate scatter plots at the top and left show studies that were not matched and thus included in only one of the two evidence bases. The density plots above and on the right show the distribution of the effect sizes across the full evidence base used by \autoagent and the human benchmark, respectively. 
    \textbf{b,}~Comparison of the pooled random-effects effect size estimated by \autoagent and the published human benchmark from Hölbling et al. (2025) \cite{Hlbling2025}. The point estimate is shown together with the corresponding 95\% confidence interval (CI).
    }
    \label{fig:effect_agreement}
\end{figure}

At the meta-analysis level, the pooled effect estimate produced by \autoagent remained close to the published benchmark, especially when accounting for the corresponding confidence intervals (Fig.~\ref{fig:effect_agreement}b). For example, for the benchmark, \autoagent estimated a positive pooled effect of Hedges' $g = 0.143$, while the original published estimate was $g = 0.020$ ($\Delta g = 0.123$; Table~\ref{tab:meta_comparison}); however, both estimates overlapped to a large extent, thus suggesting broadly similar quantitative conclusions. Importantly, the difference is within a $\pm 0.20$ tolerance region in terms of the standardized effect size estimate, which is a tolerance region that was used previously to assess whether deviations in reanalyses should be interpreted as broadly similar conclusions \cite{Aczel2026}. Both the between-study heterogeneity analysis and the publication bias diagnostics were also broadly comparable between the automated and human syntheses: both found substantial between-study heterogeneity ($I^2 = 88.3\%$ vs. 75.97\%) and evidence of small-study effects based on Egger’s test ($p = 0.006$ vs. $p = 0.018$; Table~{\ref{tab:meta_comparison}}).

At the study level, seven effect sizes could be matched between \autoagent and the human benchmark. For these matched estimates, the standardized effect sizes extracted by \autoagent showed a moderate-to-strong, positive association with the corresponding estimates from the human benchmark (Pearson’s correlation coefficient $r = 0.69$, two-sided $p = 0.085$; ordinary least squares (OLS) slope {$\beta = 1.62 $}; Fig.~\ref{fig:effect_agreement}a). This suggests that \autoagent recovered broadly similar study-level evidence. Some effect sizes were included only in one of the two evidence bases (Fig.~\ref{fig:effect_agreement}b) left and bottom inset, which reflects differences in study retrieval and eligibility assessment rather than effect size computation. For the actual quantitative synthesis, the computed effect size estimates and the resulting meta-analytic conclusions were largely similar across the automated and human evidence synthesis.

\newpage

\section{Discussion}


In this work, we introduced \autoagent, an end-to-end multi-agent framework that automates the complete meta-analysis workflow, from literature retrieval and study screening to quantitative data extraction, effect size computation, and statistical synthesis. In our evaluation, \autoagent successfully recovered 71.4\% of the studies included in the corresponding human-conducted benchmark. Furthermore, the pooled effect estimate deviated by only 0.12 Hedges’ $g$ from the published human benchmark, demonstrating that \autoagent can closely approximate expert-conducted meta-analyses. This degree of deviation is broadly consistent with human reanalysis efforts, where deviations from the original estimate are assessed using tolerance regions and where a tolerance region of $\pm 0.20$ in terms of Cohen’s $d$ is still considered as broadly acceptable \cite{Aczel2026}. The observed differences between \autoagent and the published benchmark primarily originated from literature retrieval rather than downstream statistical analysis. In particular, the largest discrepancies are because some relevant studies were not included for screening or because their full-text articles were unavailable for subsequent analysis.


Our work also contributes to the emerging field of AI-assisted scientific research. Previous LLM-based systems have demonstrated promising capabilities for literature retrieval, screening, and eligibility assessment \cite{Syriani2024,Guo2024,Boateng2024}, systematic review automation \cite{Chen2026}, and structured information extraction from scientific articles or other documents \cite{Forster2026,ahad2024,Li2026,Rouzrokh2025,xu2025,Padarha2026}. However, existing systems stop \emph{before} computing and substantially pooling standardized effect sizes across the evidence base and therefore do \emph{not} offer a quantitative synthesis. In contrast, \autoagent focuses explicitly on the steps that are specific to meta-analysis, namely, extracting statistical measures, computing standardized effect sizes  (including multiple effect sizes per study where applicable), performing meta-analytic regression, conducting publication bias diagnostics and risk-of-bias assessment, and generating reports aligned with common reporting standards.


Importantly, we do not view \autoagent as a replacement for expert reviewers. The human element remains integral to many stages of the meta-analysis workflow, such as formulating the research question, specifying inclusion and exclusion criteria, selecting relevant moderators for analysis, and interpreting findings in light of study quality and existing domain knowledge. Many of these steps require substantive expert judgment, and they can be challenging even for trained human reviewers. \autoagent should therefore not be used to shortcut expert judgment but rather to support systematic checking and to refine intermediate decisions. As such, a particularly promising application is to perform continuous updates of existing high-quality meta-analyses, where the methodological framework has already been defined, successfully implemented, and validated through peer review, so that newly published studies can be incorporated with substantially reduced manual effort. This extends the broader idea of ``living'' systematic reviews \cite{Iversen2026} to ``living'' meta-analysis, in which evidence is continuously synthesized.


The proposed \autoagent framework has several strengths. First, \autoagent covers the entire meta-analysis workflow and includes several analyses that are not always performed by default, such as publication bias diagnostics, risk-of-bias assessment, heterogeneity analyses, etc. Importantly, many studies report more than one effect size (e.g., from multiple experiments, or mixed-method approaches); \autoagent is designed to handle this feature common to meta-analyses by identifying and extracting multiple effect sizes from the same study where applicable. Second, \autoagent combines LLM-based reasoning with principled statistical procedures. Hence, the framework preserves methodological rigor while improving reproducibility and transparency. This may be especially useful for researchers with less specialized expertise in meta-analytic methods. Third, the design is modular, and, because of this, \autoagent can be used to update meta-analyses, where parts of the review protocol or evidence base may already be available. Fourth, \autoagent records traceable audit logs and intermediate outputs, including screening decisions, eligibility judgments, extracted statistics, and effect size calculations. Fifth, \autoagent is flexible with respect to the underlying LLMs. Our evaluations indicate that the framework can operate across different LLM backbones, including open-weight models, which can help improve transparency and reproducibility \cite{ Shrestha2023}.


Nevertheless, several limitations remain. First, our evaluation is limited by the number of benchmark meta-analyses. Hence, although the results demonstrate the feasibility of automated meta-analysis, they cannot establish robust performance across all potential settings, including further scientific domains, study designs, outcome types, and reporting conventions. This limitation is common for agentic systems that support scientific workflows, where comprehensive evaluation across diverse task settings is often constrained by the cost and availability of suitable benchmarks \cite{Holtdirk2026, Brodeur2026, Miao2026, Bertran2026}. It is therefore likely that performance may be lower in especially complex domains, for poorly reported studies, or for tasks requiring substantial field-specific judgment. Second, automated literature retrieval is constrained by the accessibility of scientific articles and the availability of machine-readable full text (e.g., relevant studies may be missed when full texts are behind paywalls, when PDFs cannot be reliably parsed, or when relevant statistics are reported only in figures, appendices, or other formats that are difficult to extract). Similar constraints also affect human reviewers, although humans can sometimes overcome them through institutional access or by contacting authors directly. This limitation further highlights the importance of open-access publishing and machine-readable reporting practices for scalable evidence synthesis. Third, some studies do not report sufficient quantitative information to compute standardized effect sizes. This is again a challenge shared with human meta-analysis, where reviewers must often contact authors or exclude otherwise relevant studies because necessary statistics are unavailable. Fourth, the performance of \autoagent may depend on modeling choices, including the underlying LLM and prompting strategy. Nevertheless, the modular design of \autoagent allows to compare different LLM backbones, and our evaluations further suggest that the main results are robust across different backbones, including open-weight LLMs. Fifth, training data contamination is a potential concern when evaluating LLM-based systems on published scientific studies. Because some benchmark articles may have been used for LLM training, prior exposure could, in principle, improve document interpretation, statistical extraction, or reasoning over reported results. In our setting, the risk that training data exposure directly determines the final meta-analytic result is small and mitigated in part by the sequential workflow, where effect sizes are computed through deterministic statistical procedures.  Finally, \autoagent covers many standard use cases in quantitative evidence synthesis; still, extensions may be necessary to account for field-specific differences in meta-analytic workflows.


Overall, agentic LLM systems can accelerate quantitative evidence synthesis, which is a cornerstone of scientific knowledge production and policy-making \cite{Gurevitch2018,Page2021,Higgins2024,Tyler2023}. To this end, systems such as \autoagent can help researchers, clinicians, policy-makers, and other stakeholders access reliable summaries of the rapidly growing scientific literature and thereby broaden access to rigorous evidence-based decisions.

\section{Methods}

\subsection{\autoagent}

\textbf{Overview.}
\autoagent is an end-to-end multi-agent framework for automated meta-analysis (Fig.~\ref{fig:overview}). The framework is designed to translate established best-practice principles in evidence synthesis into a structured computational workflow by mapping individual steps of a meta-analysis \cite{Page2021} to specialized agents. Starting from a research question in natural language, \autoagent first uses a \emph{planning agent} to define the meta-analysis objective, treatment and control conditions, eligibility criteria, target outcomes, candidate moderators, search strategy, and expected effect direction (using human input where needed or desired). This gives a protocol for the downstream agents that serves as a shared reference point and ensures that each stage of the workflow remains aligned with the intended meta-analysis. Based on this plan, \autoagent executes the complete evidence synthesis process, including literature retrieval (\emph{search agent}), full-text retrieval and parsing (\emph{paper reading agent}), full-text eligibility assessment (\emph{eligibility agent}), study and outcome identification (\emph{research question identification agent}), statistical data extraction (\emph{statistical result extractor agent}), statistical validation (\emph{statistical validation agent}), 
effect size computation (\emph{analysis agent}), random-effects meta-analysis (\emph{random-effects meta-analysis agent}), publication-bias assessment (\emph{bias assessment agent}), and report generation (\emph{report agent}). An overview of the different agents inside \autoagent is provided in  Table{\ref{tab:agents}}.

We implemented the LLM agents in \autoagent using the LangGraph \cite{Chase2022} framework, which executes a ReAct-style \cite{Yao2022} loop of reasoning, tool use, and observation. Here, LLM-based agents handle reasoning-intensive tasks such as document understanding and structured extraction, while deterministic computational modules perform statistical calculations, the meta-analytic regression estimation, and all visualizations. This was done to improve the methodological rigor by ensuring that statistical procedures follow established meta-analytic practice, while also making it easier to generate standardized, publication-quality visualizations with a consistent design across analyses. The prompts for the LLM-based agents are reported in the Supplementary Materials. The modular, multi-agentic design also allows users to start the workflow at later stages, for example, when updating an existing meta-analysis for which parts of the review process have already been completed, or to override specific design choices of \autoagent. All intermediate decisions and outputs are stored in a shared workflow state, which yields an auditable record of the complete meta-analysis process.

\textbf{Protocol.}
The \textit{planning agent} translates a research question in natural language into a structured meta-analysis protocol that guides the entire evidence synthesis workflow and serves as a shared representation for downstream agents. In line with established best-practice recommendations for systematic reviews and meta-analyses \cite{Page2021,Higgins2024}, the agent is prompted to prospectively specify the review objective, treatment and control conditions, eligibility criteria, target outcomes, and analysis plan. The agent also suggests the search scope, including a sensible start where applicable (e.g., to exclude studies published before a technological, clinical, or policy milestone are unlikely to be relevant, such as research on COVID-19 interventions before 2020). In addition, the agent proposes relevant search terms and candidate moderators that may be used for subgroup and heterogeneity analyses. Finally, the agent specifies the expected direction of the intervention effect, which supports downstream interpretation and helps ensure that effect sizes are coded with comparable signs across studies.

Of note, several of the above decisions made when formulating the protocol are inherently expert-driven and may require substantial domain knowledge. \autoagent therefore allows users to refine the automatically suggested protocol elements (e.g., search query, moderators, start date, databases) through a human-in-the-loop approach. For the empirical applications in this paper, we use the default protocol generated by the \textit{planning agent} to demonstrate the end-to-end automated workflow.

\textbf{Literature retrieval.}
Literature retrieval and full-text acquisition are handled by two agents as follows.

The \textit{search agent} receives the structured meta-analysis protocol and uses an LLM to generate a search query using Boolean operators tailored to the research question. Following common practice in systematic reviews, the query is designed around the core elements of the target comparison, including the treatment or intervention, comparator or control condition, and outcome of interest \cite{Higgins2024}. Where applicable, the \textit{search agent} also applies exclusion criteria to remove records that are unlikely to support quantitative synthesis, such as opinion pieces, editorials, non-empirical articles, studies without a comparison or control group, or papers that do not report quantitative outcomes.

The resulting query is then used to retrieve relevant papers from multiple scientific databases and open-access repositories (i.e., arXiv, Semantic Scholar, CrossRef, OSF Preprints, and PubMed). The databases are accessed through corresponding API calls. The retrieval pipeline is modular, so additional proprietary databases (e.g., Web of Science, Scopus, and IEEE Xplore) can be integrated where access is available. For each candidate paper, the search returns structured metadata, including the title, abstract, authors, publication year, DOI, source URL, and, when available, a direct PDF link. Retrieved records are aggregated and deduplicated using normalized publication titles. In the implementation, the number of records retrieved from each source is capped to limit cost, although this limit is configurable and can be increased for more exhaustive searches. These papers then undergo an initial rule-based screening based on the title and abstract using simple regular-expression filters to remove records that are clearly outside the scope of the review.

The \textit{paper reading agent} then attempts to retrieve the full text of each candidate paper from multiple scientific sources (e.g.,  arXiv, Semantic Scholar, CrossRef, OSF Preprints, and PubMed). When the full text cannot be obtained from these sources, additional retrieval strategies are applied, such as access the PDF via the direct PDF link from the \textit{search agent}, DOI-based resolution, and web search as a fallback. 

The retrieved PDF documents are converted into structured Markdown using a cascading document-parsing pipeline. The framework first applies MinerU \cite{Wang2024}, which we found empirically to yield the best extraction quality on a sample of papers from our corpus. Because no single tool reliably parses every PDF (e.g., scanned pages from historical databases, non-standard layouts, or transient API failures), the framework falls back in turn to LlamaParse \cite{Liu2022}, Mistral OCR \cite{MistralOCR2025}, and conventional PDF text extraction whenever the preceding method fails. Tables and document structure are preserved whenever possible, thereby enabling downstream agents to reason over both narrative text as well as statistical information embedded in tables. Papers for which the full text cannot be retrieved are logged and excluded from the subsequent steps.

\textbf{Eligibility assessment.}
The \textit{eligibility assessment agent} performs a full-text eligibility assessment for candidate studies. It reviews each complete full-text of each study to determine whether the study satisfies the predefined inclusion and exclusion criteria. In particular, the \textit{eligibility assessment agent} is instructed to check whether the study is an empirical study, has a treatment--control comparison, the availability of quantitative outcome measures, and is relevant to the research question. Exclusion reasons are logged so that decisions for each retained or discarded study can be traced during the final audit, which is also used to generate the PRISMA flowchart later.

\textbf{Study component identification.}
Scientific papers may report more than one independent empirical component, such as multiple experiments, separate samples, or mixed-method studies. The \textit{study identification agent} therefore compiles a list of different components for each paper based on the experimental design, participants, data collection procedure, and reported comparisons, so that an effect size is retrieved for each component. Specifically, for each component, the agent identifies the treatment and control groups, the relevant outcome variable, and the section of the article where the supporting statistical evidence is reported. The result is a structured representation that serves as an intermediate output (hereafter called ``map'') for the subsequent extraction stage. This further helps to ensure that statistical quantities are only collected only for effect sizes that are relevant to the objective of meta-analysis, while avoiding that irrelevant analyses, manipulation checks, or secondary results are extracted.

\textbf{Extraction of statistical information.}
To improve the reliability, \autoagent employs a structured approach to extract information. The reason is that directly asking an LLM to extract statistical results from a scientific article often leads to incomplete or incorrect outputs, as papers frequently contain multiple studies, hypotheses, outcome measures, and secondary analyses. To address this challenge, the \textit{statistical result extraction agent} is composed of two sequential stages. The two stages essentially separate document understanding from numerical extraction, which reduces the likelihood that irrelevant statistics are extracted.

First, the framework constructs a structured representation of the paper as a set of linked JSON records. The JSON records capture the experimental design, the separate study components, outcome variables, and the relationships between hypotheses and reported results via explicit cross-references between studies, effect sizes, and paper sections. This intermediate representation provides the contextual information required to accurately interpret the statistical evidence reported throughout the article, especially when the experimental design and the corresponding results are reported within different sections.

Second, statistical information is extracted only for the outcomes identified as relevant during the previous stage. For each selected result, the system records the treatment and control groups, sample sizes, descriptive statistics, test statistics, effect size information, and additional metadata required for subsequent quantitative synthesis.

\textbf{Statistical validation and relevance filtering.}
Following statistical extraction, all extracted results undergo a validation stage before quantitative synthesis. The \textit{validation agent} double-checks each extracted statistical value against the source article. The corresponding prompt instructs the agent to verify that the corresponding information can be found verbatim, or within rounding, in the relevant results section, table, or figure caption. A result is marked as ``{hallucinated}'' when core statistics (i.e., test statistic, $p$-value, or means) cannot be located anywhere in the paper text. Values that are incorrect but locatable elsewhere in the text are corrected; values that cannot be verified at all are removed. As a result, the verification step reduces hallucinated statistics and prevents incorrect values from propagating to the meta-analysis stage.

Scientific articles often report multiple statistical results for the same study, including primary outcomes, secondary analyses, subgroup analyses, and manipulation checks. To ensure that only the most relevant result contributes to the meta-analysis, \autoagent further applies a relevance filtering after statistical validation. Each extracted result is classified according to the outcome type and relevance to the research question, after which only the appropriate statistical results for quantitative synthesis are selected.

\textbf{Analysis agent.}
The \textit{analysis agent} first computes standardized effect sizes for each study component from the extracted statistical information. Here, multiple reporting formats commonly encountered in empirical studies, including hypothesis tests, regression coefficients, standardized effect measures, and odds ratios, are supported. Whenever possible, these statistics are converted into Cohen's $d$ and subsequently corrected to Hedges' $g$ \cite{Hedges1981}, providing a common effect size metric across included studies. If sufficient information is unavailable to compute a valid effect size, the corresponding result is excluded from the quantitative synthesis, and the decision is logged for traceability. 


The \textit{analysis agent} then follows a deterministic procedure to synthesize the standardized effect sizes obtained from the previous stages. Specifically, the \textit{analysis agent} estimates random-effects meta-analytic models \cite{Higgins2024} to account for the fact that the underlying studies can span different populations, interventions, outcome measures, and experimental settings and thus to account for the between-study heterogeneity. Restricted Maximum Likelihood (REML) \cite{Viechtbauer2005}. Pooled effects are reported as Hedges' $g$ \cite{Hedges1981} with corresponding 95\% confidence intervals (CIs). The random-effects meta-analytic models are estimated using a custom REML implementation.


The \textit{analysis agent} further generates publication-quality visualizations, namely, (i) forest plots and (ii)~cumulative forest plots across years to analyze how the pooled effect size changes as new evidence is accumulated over time. In addition, the \textit{analysis agent} computes standard heterogeneity statistics, including Cochran's $Q$ \cite{Deeks2019}, $I^2$, and $\tau^2$, to quantify variability across studies \cite{Borenstein2009}. Where sufficient information is available, the agent also conducts moderator analyses to examine whether effect sizes vary systematically across predefined moderators such as specific study characteristics (e.g., the choice of the treatment).

\textbf{Bias assessment.}
The \textit{bias assessment agent} evaluates potential publication bias and study-level risk of bias. For publication bias, the agent applies established methods when a sufficient number of studies are available. Specifically, it assesses potential bias using three complementary analyses: (i) Egger's regression test \cite{Egger1997} to test for small-study effects; (ii) the trim-and-fill procedure \cite{Duval2000} to estimate the potential impact of missing studies on the pooled effect size; (iii) leave-one-out sensitivity analysis to examine whether the pooled estimate depends strongly on any individual study; and (iv)~a scatter plot of effect sizes against sample size as an exploratory diagnostic for small-study effects and potential publication bias. The final report includes (i)--(iv). The agent also generates funnel plots to support visual inspection of small-study effects. Random-effects meta-analytic models were estimated using a custom implementation of Restricted Maximum Likelihood (REML) in Python (NumPy/SciPy), following the iterative algorithm of Viechtbauer \cite{Viechtbauer2005}. The DerSimonian--Laird estimator \cite{Jackson2010} was used to initialize the between-study variance ($\tau^2$) before REML optimization. Small-study effects and potential publication bias were assessed using custom implementations of Egger's regression test \cite{Egger1997} and the trim-and-fill procedure \cite{Duval2000}. Rather than relying on external meta-analysis libraries (e.g., \texttt{metafor} or \texttt{statsmodels}), all statistical routines were implemented directly in Python to ensure full control over the analysis pipeline and facilitate integration with the multi-agent framework.

For study-level risk of bias, the \textit{Bias assessment agent} uses the assessment framework appropriate to the study design, such as RoB~2 for randomized trials \cite{Sterne2019} and ROBINS-I for non-randomized studies of interventions \cite{Sterne2016}. The resulting risk-of-bias judgments are reported in structured tabular form.

\textbf{Report.} 
Upon completion of the workflow, the \textit{report agent} automatically generates a structured report that summarizes the full evidence synthesis. The report includes the literature retrieval process, study selection decisions, extracted quantitative information, effect size calculations, random-effects meta-analysis, moderator analyses to assess heterogeneity, publication bias assessments, risk-of-bias assessment, and all corresponding visualizations. The \textit{report agent} also produces a PRISMA 2020 flow diagram \cite{Page2021} that reports the number of study records in each stage of the review, including exclusion reasons and the final number of effect sizes. In addition, it generates a structured summary table containing all included studies and key metadata, such as study identifier, sample size, intervention or exposure, outcome, and the extracted effect size. All plots are generated using the matplotlib plotting library and exported as static PNG/PDF figures.

To improve transparency, \autoagent records the outputs of every framework stage, including screening decisions, extracted statistics, effect size calculations, etc. These records allow users to trace and audit each included or excluded study throughout the workflow and facilitate a human-in-the-loop use of \autoagent to refine, audit, extend, or update existing meta-analyses.

\subsection{Evaluation}


To evaluate the performance of \autoagent, we benchmark our framework against published manual meta-analyses. Here, we use the meta-analysis by Hölbling et al. (2025) \cite{Hlbling2025} with $k=12$ effect sizes from $N=17,422$ participants as our main benchmark. First, it is a peer-reviewed and recent meta-analysis, making it a timely reference case for which the underlying publications are available online, so that future work has a transparent basis to benchmark \autoagent. Second, it includes a broad range of analyses expected in a principled, state-of-the-art meta-analysis to test the full capability spectrum of \autoagent. For example, it includes multiple heterogeneity analyses to demonstrate how \autoagent can analyze moderators. Third, key information of the meta-analysis workflow in \cite{Hlbling2025} is public, including the structured extraction tables. For example, it also reports risk-of-bias assessments in structured numerical tabular form to facilitate quantitative comparisons across human approaches and \autoagent. Finally, because one author of the present study overlaps with the original lead of the review team, we can complement the quantitative benchmark with a qualitative assessment of where \autoagent agrees with, diverges from, or misses decisions made in the human benchmark. 

The input to \autoagent was the following research question (see Fig.~\ref{fig:dashboard}a): ``\emph{What is the effect of large language model (LLM)-generated persuasive messages on persuasive outcomes compared with non-LLM or human-authored control messages?}'' The corresponding search query generated by \autoagent is shown in (Fig.~\ref{fig:dashboard}a). We did not specify a start or end date for the search; the search was performed in July 2026, so all studies up to this date were included as knowledge cut-off.


We evaluate \autoagent in two ways. First, we assess the intermediate performance by validating different steps of the evidence synthesis workflow by comparing the outputs of individual agents with the human benchmark. (i)~For the \emph{search agent}, we evaluate literature retrieval 
using overlap, precision, and recall between studies automatically retrieved by \autoagent vs. the studies retrieved or included in the benchmark. (ii)~For the \emph{eligibility agents}, we compare inclusion and exclusion decisions using confusion matrices. (iii)~For the \emph{statistical extraction agent} and \emph{validation agent}, we compare the extracted quantitative information, including effect sizes and standard errors, against the human benchmark.


Second, we evaluate the final quantitative synthesis produced by \autoagent. We compare the pooled random-effects Hedges' $g$ estimate and the confidence interval with the corresponding estimate reported in the human benchmark. Effect size agreement is summarized graphically using scatter plots of benchmark against \autoagent estimates. We report Pearson's correlation coefficient $r$ to quantify the association between the two sets of effect sizes. In addition, to quantify the calibration, we estimate the slope $\beta$ from an ordinary least squares (OLS) regression with an intercept, where we regress the benchmark effect size on the corresponding \autoagent effect size.

\subsection{Implementation details}


\autoagent is implemented in Python 3.11 and uses LangGraph \cite{Chase2022} to orchestrate the different agents. All LLM-based agents (i.e., the \emph{planning agent}, \emph{search agent}, \emph{eligibility agent}, \emph{research question identification agent}, \emph{statistical result extractor agent}, \emph{statistical validation agent}, \emph{relevance filter agent}, and \emph{report agent}) use  \texttt{(openai/gpt-5.4-mini)} through the OpenRouter API \cite{openrouter2026}. 


Reporting of the LLM configuration follows best-practice recommendations for LLM-based research \cite{Feuerriegel2026}; further details, prompts, and configuration files are provided in the repository. 
We set the native reasoning effort to \emph{medium}, 
we limited web tool use for retrieval to $40$ papers extracted/database, and we left all other decoding parameters at their provider defaults. We do not use persistent conversational memory between agent calls; each agent receives only the relevant workflow state and source material required for its task. Structured outputs using JSON format are used where required, for example, for eligibility decisions, extracted statistical quantities, and the statistical validation. 



A complete end-to-end run of main analysis using the H{\"o}lbling et al. benchmark \cite{Hlbling2025} took $\sim$0.5 hours and used approximately 1M input tokens and 100K output tokens across 200 LLM calls. This corresponded to a cost of around \$1.5 under the model pricing at the time of access.


During the early development of \autoagent, we explored simpler alternatives, including a single-agent workflow and an integrated commercial agentic system for scientific workflows (i.e., Claude Science \cite{Anthropic2025}). These approaches were useful for isolated subtasks but did not reliably complete the full evidence synthesis workflow; failures often appeared during long-running executions such as study retrieval and quantitative extraction. As a result, the end-to-end performance was not sufficiently stable for a meaningful benchmark, and we thus used the experiments primarily to motivate the proposed multi-agent framework, which maps the manual steps of literature reviews and meta-analyses onto a combination of specialized agents with deterministic statistical processing.

\vspace{0.4cm}
\section*{Data availability}

All data needed to replicate our analyses will be made available via a dedicated, public GitHub repository.

\vspace{0.4cm}
\section*{Code availability}

\autoagent will be made publicly available under an open-source license.


\newpage

\bibliography{literature} 


\newpage
\section*{Acknowledgments}

Funding by the Deutsche Forschungsgemeinschaft (DFG, German Research Foundation) under the National Research Data Infrastructure -- NFDI 27/1-2026, project number 460037581 is acknowledged.

\vspace{0.4cm}
\section*{Author contributions} 

All authors contributed to conceptualization, manuscript writing, and approved the manuscript.

\vspace{0.4cm}
\section*{Competing interests}
The authors declare no competing interests.

\newpage

\appendix

\setcounter{figure}{0}
\setcounter{table}{0}

\renewcommand{\thefigure}{S\arabic{figure}}
\renewcommand{\thetable}{S\arabic{table}}

\renewcommand{\figurename}{Fig.}
\renewcommand{\tablename}{Table}

\section*{Supplementary Figures}

\begin{figure}[H]
    \centering
    \includegraphics[width=0.65\linewidth]{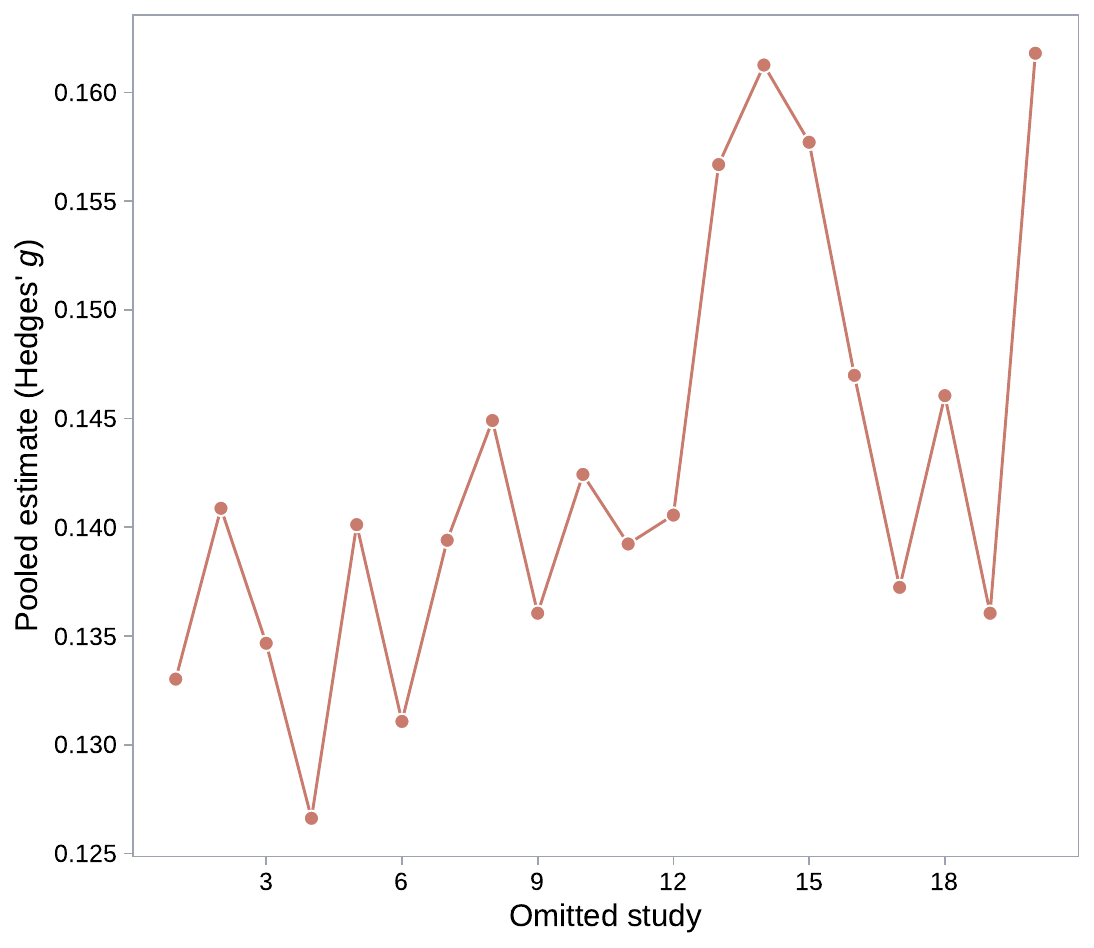}
    \caption{
        {\textbf{Leave-one-out sensitivity analysis.} The pooled random-effects estimate (Hedges' $g$) was recomputed $k$ times, each time omitting one study, to assess whether any single study disproportionately drives the overall result. Each point shows the recomputed pooled estimate with that study excluded, plotted against the omitted study's index; the estimate remains stable across omissions (range $g \approx 0.13$--$0.16$), indicating that no single study is unduly influential.}}
    \label{fig:leave_one_out}
\end{figure}

\newpage

\section*{Supplementary Tables}

\subsection*{Moderator analysis}

\begin{table}[H]
\centering
\caption{Subgroup (moderator) analyses of the pooled effect size (Hedges' $g$). Subgroups with fewer than two studies were not pooled; these include Country (Canada, United Kingdom), LLM model (Bloom, GPT-3 davinci, GPT-4o, GPT-4-0613), and Publication status (Working paper).}
\label{tab:moderator_analysis}

\begin{tabular}{llrrrrrr}
\toprule
Moderator & Subgroup & $k$ & Hedges' $g$ & 95\% CI & $I^2$ (\%) & $Q_M$ & $p$ \\
\midrule

\multirow{3}{*}{Study design}
 & Between subjects & 15 & $0.147$ & $[0.054,\,0.239]$ & 89.4 & \multirow{3}{*}{$0.87$} & \multirow{3}{*}{.646} \\
 & Mixed & 2 & $-0.043$ & $[-0.796,\,0.709]$ & 95.4 &  &  \\
 & Within subjects & 3 & $0.184$ & $[0.088,\,0.280]$ & 66.0 &  &  \\
\midrule

Country
 & United States & 19 & $0.147$ & $[0.060,\,0.234]$ & 88.8 & $0.22$ & .642 \\
\midrule

\multirow{4}{*}{LLM model}
 & Claude 3.5 Sonnet & 2 & $0.393$ & $[0.198,\,0.589]$ & 64.0 & \multirow{4}{*}{$13.16$} & \multirow{4}{*}{.068} \\
 & DeepSeek v3 & 2 & $0.282$ & $[0.110,\,0.454]$ & 66.6 &  &  \\
 & GPT-3 & 8 & $0.067$ & $[-0.055,\,0.189]$ & 89.6 &  &  \\
 & GPT-3.5 & 4 & $0.173$ & $[0.076,\,0.269]$ & 76.8 &  &  \\
\midrule

\multirow{3}{*}{Domain}
 & General persuasion & 5 & $0.331$ & $[0.243,\,0.418]$ & 44.2 & \multirow{3}{*}{$10.88$} & \multirow{3}{*}{.004} \\
 & Health & 8 & $-0.018$ & $[-0.153,\,0.116]$ & 85.9 &  &  \\
 & Politics & 7 & $0.182$ & $[0.128,\,0.236]$ & 59.0 &  &  \\
\midrule

Publication status
 & Published & 19 & $0.133$ & $[0.048,\,0.219]$ & 88.3 & $1.00$ & .318 \\
\bottomrule
\end{tabular}

\end{table}

\begin{table}[H]
\centering
\caption{\textbf{Comparison of quantitative evidence synthesis for \autoagent and the human bechmark.} Results are based on the LLM persuasion task from Hölbling et al. (2025) \cite{Hlbling2025}.}
\label{tab:meta_comparison}

\small
\begin{tabular}{lccc}
\toprule
\textbf{} &
\textbf{Manual benchmark} &
\textbf{\autoagent} \\
\midrule

Included studies          & 7      & 8    \\
Effect size estimates ($k$)          & 12     & 20    \\

\midrule

Pooled hedges' $g$        & 0.020  & 0.143  \\
95\% CI                   & [-0.048, 0.093]   & [0.059, 0.226] \\
$p$-value                 & 0.530  & $<0.0008$ \\

\midrule

$I^2$ (\%)                & 75.97  & 88.3  \\
Egger's test ($p$-value)        & 0.018  & 0.006  \\

\bottomrule
\end{tabular}
\end{table}

\subsection*{Overview of Multi-Agent Architecture}

\begin{table}[H]
\singlespacing
\centering
\scriptsize
\setlength{\tabcolsep}{3.5pt}
\renewcommand{\arraystretch}{1.10}

\caption{Overview of the different agents within \textsc{\autoagent}.}
\label{tab:agents}

\begin{tabular}{@{}p{2.0cm}p{4.9cm}p{2.5cm}p{2.7cm}p{1.1cm}@{}}
\toprule
\textbf{Agent} & \textbf{Purpose} & \textbf{Input} & \textbf{Output} & \textbf{Prompt} \\
\midrule

Planning &
Translates the natural-language research question into a structured review protocol &
Research question &
Meta-analysis plan, search queries, inclusion criteria, moderators &
\ref{prompt:planning} \\
\midrule

Search &
Generates and reformulates search queries, retrieves and deduplicates candidate studies, applies title/abstract screening &
Meta-analysis plan &
List of candidate papers (with metadata) &
\ref{prompt:search} \\
\midrule

Paper reading &
Retrieves and parses full text via a document extraction pipeline &
Candidate paper metadata / PDFs &
Full-text papers (structured Markdown) &
N/A \\
\midrule

Eligibility &
Assesses full-text eligibility against inclusion/exclusion criteria &
Full-text papers &
Eligible papers, exclusion reasons &
\ref{prompt:eligibility} \\
\midrule

Research question identification &
Identifies independent research questions within each paper and maps relevant research questions/outcomes &
Eligible full-text papers &
Studies and their relevance to the search queries &
\ref{prompt:rq_indentification} \\
\midrule

Statistical result extractor &
Builds a structural map of the paper, then extracts quantitative results for each relevant outcome &
Structured paper representation, full text &
Extracted statistical results &
\ref{prompt:extractor} \\
\midrule

Statistical validation &
Validates extracted values against the source text; corrects or removes unsupported quantitative evidence &
Extracted statistical results, full text &
Validated statistical results &
\ref{prompt:checker} \\
\midrule

Relevance filter &
Classifies extracted rows and selects a single winning result per study--outcome group &
Verified result rows &
Filtered rows (one per outcome) &
\ref{prompt:relevance_filter} \\
\midrule

Analysis agent &
Computes standardized effect sizes (Hedges' $g$, Cohen's $d$) and synthesizes them via a random-effects model &
Filtered, coded results &
Pooled effect size, heterogeneity ($I^2$, $\tau^2$), plots &
N/A \\
\midrule

Random-effects meta-analysis &
Computes standardized effect sizes (Hedges' $g$), estimates between-study heterogeneity using REML, and synthesizes evidence using a random-effects meta-analysis &
Validated statistical results &
Pooled effect size, heterogeneity statistics ($Q$, $I^2$, $\tau^2$), confidence intervals, and meta-analysis visualizations &
N/A \\
\midrule

Bias assessment &
Assesses publication bias and study-level risk of bias using established meta-analysis methods &
Included studies, extracted statistical results, pooled effect sizes &
RoB assessment, Egger's test, trim-and-fill analysis, funnel plot &
\ref{prompt:bias} \\
\midrule

Report &
Synthesizes all prior outputs into a PRISMA-compliant narrative report &
All prior-stage outputs &
Final meta-analysis report &
\ref{prompt:report} \\

\bottomrule
\end{tabular}
\end{table}

\newpage

\section*{Supplementary Materials}
\label{supp:supplementary_materials}

\subsection*{Prompts}

\singlespacing

\subsubsection*{Planning agent}
\phantomsection
\label{prompt:planning}
\begin{tcolorbox}[breakable, colback=gray!5, width=\textwidth, colframe=gray!80, title=Planning agent: System prompt]
\footnotesize
\begin{verbatim}
You are an expert meta-analyst and systematic review methodologist. A user
has given you a research question. Your job is to design a complete
meta-analysis plan for that question following PRISMA 2020 guidelines.

You must call `create_plan` with ALL fields filled in. Be specific:
  - treatment_label / control_label: the two conditions being compared
    (e.g. 'LLM-generated persuasive text' vs 'human-written persuasive
    text').
  - search_queries: 2-3 short keyword strings optimised for academic
    database search (not Boolean syntax -- just keyword phrases). Each
    query should include the intervention/topic plus empirical intent
    words when natural, such as 'experiment', 'trial', 'controlled',
    'comparison', 'quantitative', 'outcome', or 'effect'. Avoid queries
    that mainly retrieve reviews or meta-analyses.
  - inclusion_criteria: 2-3 BROAD, permissive criteria. Keep them
    general -- err heavily on the side of inclusion. A paper only needs
    to be broadly related to the topic and report some quantitative
    outcome. Do NOT add strict design, sample-size, or journal-quality
    requirements.
  - exclusion_criteria: 3-4 exclusion reasons that MUST include:
    (1) not an original empirical study (systematic reviews,
        meta-analyses, editorials, opinion pieces, and theoretical
        papers are excluded);
    (2) no comparison between a treatment and a control condition;
    (3) no quantitative outcome data that can produce an effect size
        (means+SDs, t/F-values, proportions, or odds ratios required);
    (4) completely unrelated to the research topic.
  - moderators_to_extract: 3-6 study-level variables that could explain
    heterogeneity (e.g. 'study design', 'topic domain', 'sample size
    category').
  - positive_direction: what a POSITIVE Hedges g means in plain
    language.
  - search_start_year: sensible earliest year (e.g. 2015 for LLM
    topics, 1990 for clinical trials).
\end{verbatim}
\end{tcolorbox}

\subsubsection*{Search agent}
\phantomsection
\label{prompt:search}
\begin{tcolorbox}[breakable, colback=gray!5, width=\textwidth, colframe=gray!80, title=Search agent: Query generation prompt]
\footnotesize
\begin{verbatim}
You are helping run a systematic literature search for a meta-analysis.
Rewrite the planned research question into concise academic database
search queries.
Return exactly {max_variants} query string(s), unless fewer are truly
appropriate.

Research question: {research_question}
Treatment/intervention: {treatment_label}
Control/comparator: {control_label}
Outcome: {outcome_description}
Existing draft queries: {existing_draft_queries}

Guidelines:
- Use short keyword phrases, not long sentences.
- Include empirical intent words when natural, such as experiment,
  trial, controlled, comparison, quantitative, outcome, or effect.
- Avoid phrases that mainly retrieve reviews, systematic reviews,
  meta-analyses, editorials, protocols, or opinion pieces.
- Do not use Boolean operators unless they are essential.

Return only JSON in this schema: {"queries": ["query one", "query two"]}
\end{verbatim}
\end{tcolorbox}

\subsubsection*{Eligibility agent}
\phantomsection
\label{prompt:eligibility}
\begin{tcolorbox}[breakable, colback=gray!5, width=\textwidth, colframe=gray!80, title=Eligibility agent: System prompt (templated per plan)]
\footnotesize
\begin{verbatim}
You are a systematic review methodologist assessing papers for inclusion
in a quantitative meta-analysis.

Research question: {research_question}

Inclusion criteria:
{inclusion_criteria}

Exclusion criteria:
{exclusion_criteria}

Only EXCLUDE the paper when the evidence is clear. Do not infer
exclusion from missing snippets, sparse abstracts, or uncertain
terminology.

EXCLUDE the paper if ANY of the following is clearly true:
  A) It is NOT an original empirical study -- it is a systematic
     review, meta-analysis, editorial, commentary, opinion piece, or
     purely theoretical paper.
  B) It has NO comparison between at least two groups or conditions
     (no treatment vs. control, intervention vs. baseline, or
     equivalent contrast).
  C) It clearly has NO quantitative outcome data -- no means, SDs,
     t-values, F-values, proportions, or odds ratios that could
     support an effect size computation.
  D) It is completely unrelated to the research question.

INCLUDE if none of the above clearly apply. When in doubt about
design, topic fit, or quantitative data availability, INCLUDE -- the
extractor and deterministic statistical checks will verify this
downstream.

Call `eligibility_decision` for every paper. Always provide a short
reason (<=8 words), e.g. 'systematic review', 'no control condition',
'no quantitative outcome data', 'meets all criteria'.
\end{verbatim}
\end{tcolorbox}

\subsubsection*{Research question identification agent}
\phantomsection
\label{prompt:rq_indentification}

\begin{tcolorbox}[breakable, colback=gray!5, width=\textwidth, colframe=gray!80, title=Study identification agent: System prompt]
\footnotesize
\begin{verbatim}
You are a scientific-paper structure analyst specialised in identifying 
independent studies within a single paper.

A paper may report one study or several (Study 1 / Study 2, Experiment 1 / 
Experiment 2, Wave 1 / Wave 2, etc.). Your job is to detect every 
independent study and record its sample size, design, treatment/control 
groups, primary outcome variables, and where its results appear in the 
paper.

Rules:
- A 'study' is an independent data-collection episode with its own sample. 
Separate conditions within ONE dataset are NOT separate studies.
- Do not invent studies. If the paper has one study, record only one.
- Every study record must cite where in the paper its results appear.
- Call `record_study` once per study.
\end{verbatim}
\end{tcolorbox}

\begin{tcolorbox}[breakable, colback=gray!5, width=\textwidth, colframe=gray!80, title=Research question identification agent: System prompt]
\footnotesize
\begin{verbatim}
You are a scientific meta-analysis extraction specialist.

Your task is NOT to extract every statistical analysis or every hypothesis
in the paper.

Your task is to extract ONLY the experimentally tested effects that are 
directly relevant to the synthesis objective provided by the user.

The synthesis objective defines:
  * which constructs are relevant,
  * which outcomes are primary,
  * and which analyses should be ignored.

--------------------------------------
GENERAL EXTRACTION RULES
--------------------------------------
Extract:
  * primary experimentally tested effects
  * preregistered or central outcomes
  * statistically evaluated comparisons directly relevant to the synthesis 
  objective

Do NOT extract:
  * manipulation checks
  * robustness checks
  * mediation analyses
  * moderation analyses unless central
  * exploratory analyses
  * manipulation validation
  * linguistic/process variables
  * descriptive-only statistics
  * supplementary analyses
  * subgroup analyses unless central to the paper's main claim

--------------------------------------
UNIT OF EXTRACTION
--------------------------------------
The unit of extraction is:
ONE DISTINCT EFFECT RELEVANT TO THE SYNTHESIS OBJECTIVE

NOT:
  * every table row
  * every regression
  * every statistical test

If multiple analyses test the same construct:
extract only the primary/preferred/main analysis.

--------------------------------------
PRIORITIZATION
--------------------------------------
Prioritize:
  1. Primary outcomes
  2. Preregistered outcomes
  3. Main experimental contrasts
  4. Central claims in abstract/results/discussion

If uncertain whether an effect is central:
prefer inclusion — it is better to over-extract than to miss a relevant 
effect.

--------------------------------------
OUTPUT PHILOSOPHY
--------------------------------------
You are building a CLEAN SYNTHESIS DATASET,
not a complete theory map.

Fewer highly relevant effects are preferred over many weakly related 
analyses.
\end{verbatim}
\end{tcolorbox}

\subsubsection*{Statistical result extractor agent}
\phantomsection
\label{prompt:extractor}
\begin{tcolorbox}[breakable, colback=gray!5, width=\textwidth, colframe=gray!80, title=Statistical result extractor agent: System prompt]
\footnotesize
\begin{verbatim}
You are a precise research-paper analyst and statistical extractor.
You work in two mandatory phases:

PHASE 1 -- Paper Understanding (ALWAYS FIRST)
  Call `record_paper_map` ONCE before anything else.
  Build a complete structural map: paper type, study design, comparison
  type, studies, outcome variables, hypotheses (H1/H2/RQ1/...), and the
  sections where statistics appear.
  For each hypothesis record: hypothesis_id, hypothesis_text,
  linked_outcome (the outcome variable it tests), study_label, and
  predicted direction.

PHASE 2 -- Statistical Extraction
  Use the paper map from Phase 1 as your mental model.
  When a user query is provided, use the hypotheses list to identify
  ALL hypotheses whose linked_outcome directly or closely measures the
  construct in the user query -- then extract the statistical result
  for EACH relevant hypothesis in EACH independent study.
  Extract ONLY from this paper's own Results sections -- never from
  Introduction, Related Work, Background, or Discussion sections that
  describe OTHER papers' statistics.

Phase 1 must always complete before Phase 2 begins. Never skip the
paper map. Never invent experiments or statistics.
\end{verbatim}
\end{tcolorbox}

\subsubsection*{Statistical validation agent}
\phantomsection
\label{prompt:checker}
\begin{tcolorbox}[breakable, colback=gray!5, width=\textwidth, colframe=gray!80, title=Statistical validation agent: System prompt]
\footnotesize
\begin{verbatim}
You are a rigorous statistical fact-checker for meta-analysis.

You are given the full text of a research paper and a list of
statistical results that were automatically extracted from it. Your
job is to verify each result by locating the exact numbers in the
paper text.

Rules:
1. Only accept a number as verified if it appears VERBATIM (or within
   rounding) in the paper text -- in the Results section, tables, or
   figure captions.
2. Do NOT accept numbers from Introduction, Related Work, or
   Discussion sections that describe OTHER studies.
3. If a value is wrong but the correct value is in the text, issue a
   'corrected' verdict and provide the corrected value.
4. If a value cannot be located anywhere in the text, mark it in
   failed_stats.
5. Mark the result 'hallucinated' only if the CORE numbers (test
   statistic, p-value, or means) are absent from the paper text
   entirely.
6. Call `record_validation` once per extracted result row.
7. Be precise: 3.77 and 3.78 are NOT the same.
\end{verbatim}
\end{tcolorbox}

\subsubsection*{Relevance filter agent}
\phantomsection
\label{prompt:relevance_filter}
\begin{tcolorbox}[breakable, colback=gray!5, width=\textwidth, colframe=gray!80, title=Relevance filter agent: System prompt]
\footnotesize
\begin{verbatim}
You are a meta-analysis outcome classifier. Your ONLY job is to
classify each row -- you do NOT decide which rows to keep. A separate
algorithm will select winners based on your classifications.

For EVERY row call classify_and_select with accurate values for:

1. OUTCOME CLASS
   primary | secondary | manipulation_check | process_variable |
   exploratory | robustness_check | descriptive

2. CONSTRUCT FAMILY
   A normalised label for the underlying construct being measured.
   CRITICAL -- PARENTHETICAL QUALIFIER RULE:
   Outcome names that share the same base name but differ only by a
   parenthetical qualifier are sub-conditions of the SAME construct.
   Strip the qualifier.
   Examples:
     'Compliance rate (overall)', 'Compliance rate (truthful)',
     'Compliance rate (deceptive)'
     -> all three: construct_family = 'compliance_rate'
     'Accuracy (truthful persuasion)', 'Accuracy (deceptive
     persuasion)'
     -> both: construct_family = 'accuracy'
     'Attitude change (immediate)', 'Attitude change (delayed)'
     -> both: construct_family = 'attitude_change'
   The qualifier NEVER creates a new construct_family.

3. GRANULARITY LEVEL
   Within the construct_family group, is this row:
   - 'aggregate'     -- covers all conditions/full sample (keyword:
     'overall', 'total', 'combined', or no qualifier)
   - 'sub_condition' -- splits by condition, content type, time point,
     sub-scale (keywords: 'truthful', 'deceptive', 'immediate',
     'delayed', etc.)
   - 'standalone'    -- neither (single unique measure)

4. COMPARISON ARM
   From group_1_label / group_2_label, classify the comparison arm:
   - 'target_arm'     -- treatment is the LLM / AI intervention of
     interest
   - 'non_target_arm' -- treatment is a human or other non-AI
     comparator
   - 'primary'        -- only one arm exists

5. IS SYNTHESIS RELEVANT
   True if the construct directly answers the synthesis objective.
   False for manipulation checks, process variables unrelated to the
   main outcome, etc.

6. PRIORITY RANK (1 = highest)
   1=behavior, 2=compliance/persuasion, 3=attitude, 4=intention,
   5=resistance, 6=experiential, 7=mechanism, 8=descriptive

Call classify_and_select once for EVERY row. Do not skip any.
Set is_winner=True for all rows -- the selection algorithm overrides
this.
\end{verbatim}
\end{tcolorbox}

\subsubsection*{Analysis agent}
\phantomsection
\label{prompt:meta_analysis}
\begin{tcolorbox}[breakable, colback=gray!5, width=\textwidth, colframe=gray!80, title= Analysis agent: Interpretation prompt (templated per result)]
\footnotesize
\begin{verbatim}
You are a meta-analysis expert. Summarise these results in 3-4
sentences suitable for a systematic review paper:
  Pooled Hedges' g = {pooled_g}, 95% CI [{ci_lower}, {ci_upper}],
  p = {p_value}, I^2 = {i_squared}%, tau = {tau}, k = {k} studies.
Focus on effect magnitude, direction, and heterogeneity.
\end{verbatim}
\end{tcolorbox}

\subsubsection*{Bias assessment agent}
\phantomsection
\label{prompt:bias}
\begin{tcolorbox}[breakable, colback=gray!5, width=\textwidth, colframe=gray!80, title=Bias agent: Interpretation prompt (templated per result)]
\footnotesize
\begin{verbatim}
Interpret this publication bias assessment in 2-3 sentences:
Egger's test p={eggers_p_value}, trim-and-fill k0={trim_fill_k0},
risk={bias_risk}.
\end{verbatim}
\end{tcolorbox}

\subsubsection*{Report agent}
\phantomsection
\label{prompt:report}
\begin{tcolorbox}[breakable, colback=gray!5, width=\textwidth, colframe=gray!80, title=Report agent: System prompt (templated per plan)]
\footnotesize
\begin{verbatim}
You are an academic writer specialising in systematic reviews and
meta-analyses. Write a comprehensive, publication-quality meta-analysis
report.

Research question: {research_question}
Treatment: {treatment_label}
Control: {control_label}
Outcome: {outcome_description}
Effect direction: {positive_direction}

Structure: Abstract -> Search/Screening -> Included Studies -> Effect
Sizes -> Meta-Analytic Result -> Heterogeneity -> Moderators ->
Publication Bias -> Discussion -> Limitations -> Conclusion.
Call `finalise_report` with the complete Markdown text.
\end{verbatim}
\end{tcolorbox}


\begin{landscape}
\singlespacing
\scriptsize
\setlength{\tabcolsep}{4pt}
\renewcommand{\arraystretch}{1.15}

\begin{longtable}{%
>{\raggedright\arraybackslash}p{0.9cm}
>{\raggedright\arraybackslash}p{0.9cm}
>{\raggedright\arraybackslash}p{4.8cm}
>{\raggedright\arraybackslash}p{2.6cm}
>{\raggedright\arraybackslash}p{4.0cm}
>{\raggedright\arraybackslash}p{5.2cm}}
\caption{Study-level metadata for the studies included in the meta-analysis, as extracted by \autoagent.}
\label{tab:study_table_export} \\

\toprule
Study & Report & Title & Author & Year & LLM model \\
\midrule
\endfirsthead

\toprule
Study & Report & Title & Author & Year & LLM model \\
\midrule
\endhead

\bottomrule
\endfoot

S1 & R1 & On the Conversational Persuasiveness of Large Language Models: A Randomized Controlled Trial & Salvi & 2024 & gpt-4-0613 (GPT-4) \\
S3 & R1 & How persuasive is AI-generated propaganda? & Gulati & 2024 & GPT-3 davinci \\
S7 & R1 & Large Language Models Are More Persuasive Than Incentivized Human Persuaders & Schoenegger et al. & 2025 & Claude 3.5 Sonnet \\
S7 & R2 & Large Language Models Are More Persuasive Than Incentivized Human Persuaders & Schoenegger et al. & 2025 & Claude 3.5 Sonnet \\
S8 & R1 & Large Language Models Are More Persuasive Than Incentivized Human Persuaders & Schoenegger et al. & 2025 & DeepSeek v3 \\
S8 & R2 & Large Language Models Are More Persuasive Than Incentivized Human Persuaders & Schoenegger et al. & 2025 & DeepSeek v3 \\
S9 & R1 & LLM-generated messages can persuade humans on policy issues & Bai & 2025 & GPT-3 \\
S10 & R1 & LLM-generated messages can persuade humans on policy issues & Bai & 2025 & GPT-3.5 \\
S11 & R1 & LLM-generated messages can persuade humans on policy issues & Bai & 2025 & GPT-3.5 \\
S19 & R1 & Working with AI to persuade: Examining a large language model’s ability to generate pro-vaccination messages & Karinshak et al. & 2023 & GPT-3 \\
S19 & R2 & Working with AI to persuade: Examining a large language model’s ability to generate pro-vaccination messages & Karinshak et al. & 2023 & GPT-3 \\
S19 & R3 & Working with AI to persuade: Examining a large language model’s ability to generate pro-vaccination messages & Karinshak et al. & 2023 & GPT-3 \\
S20 & R1 & Working with AI to persuade: Examining a large language model’s ability to generate pro-vaccination messages & Karinshak et al. & 2023 & GPT-3 \\
S20 & R2 & Working with AI to persuade: Examining a large language model’s ability to generate pro-vaccination messages & Karinshak et al. & 2023 & GPT-3 \\
S20 & R3 & Working with AI to persuade: Examining a large language model’s ability to generate pro-vaccination messages & Karinshak et al. & 2023 & GPT-3 \\
S21 & R1 & Conversations with AI Chatbots Increase Short-Term Vaccine Intentions But Do Not Outperform Standard Public Health Messaging & Sehgal & 2025 & GPT-4o \\
S22 & R1 & AI-Generated Messages Can Be Used to Persuade Humans on Policy Issues & Bai. & 2025 & GPT-3 \\
S23 & R1 & AI-Generated Messages Can Be Used to Persuade Humans on Policy Issues & Bai. & 2025 & GPT-3.5 \\
S24 & R1 & AI-Generated Messages Can Be Used to Persuade Humans on Policy Issues & Bai. & 2025 & GPT-3.5 \\
S26 & R1 & The effect of source disclosure on evaluation of AI-generated messages: A two-part study & Lim & 2023 & Bloom \\
\end{longtable}

\begin{longtable}{%
>{\raggedright\arraybackslash}p{0.9cm}
>{\raggedright\arraybackslash}p{2.8cm}
>{\raggedright\arraybackslash}p{4.0cm}
>{\raggedright\arraybackslash}p{3.6cm}
>{\raggedright\arraybackslash}p{4.0cm}
>{\raggedright\arraybackslash}p{4.8cm}}
\caption{Study and outcome characteristics for the same meta-analytic corpus.}
\label{tab:outcome_table_export} \\

\toprule
Study & Domain & Interaction type & Recruitment source & Outcome type & Outcome measurement \\
\midrule
\endfirsthead

\toprule
Study & Domain & Interaction type & Recruitment source & Outcome type & Outcome measurement \\
\midrule
\endhead

\bottomrule
\endfoot

S1 & conversational persuasion / online debates & one-on-one debate & Prolific & ordinal & 1--5 Likert transformed agreement outcome analyzed with partial proportional odds model \\
S3 & political persuasion / propaganda & one-shot article exposure & Lucid quota sample & percent agreement & percentage agreement with thesis statements \\
S7 & persuasion / decision-making & interactive conversational persuasion & Prolific & behavioral persuasion/compliance & average compliance rate across persuasion questions \\
S7 & persuasion / decision-making & interactive conversational persuasion & Prolific & behavioral accuracy & percentage of correct answers \\
S8 & persuasion / decision-making & interactive conversational persuasion & Prolific & behavioral persuasion/compliance & average compliance rate across deceptive persuasion questions \\
S8 & persuasion / decision-making & interactive conversational persuasion & Prolific & behavioral accuracy & percentage of correct answers \\
S9 & political persuasion / policy attitudes & message exposure & Prolific.com & attitude change & policy support for a smoking ban \\
S10 & political persuasion / policy attitudes & message exposure & Prolific.com and CloudResearch & attitude change & policy support for an assault weapons ban \\
S11 & political persuasion / policy attitudes & message exposure & Prolific.com & attitude change & policy support across policy topics \\
S19 & public health messaging & message exposure & Amazon Mechanical Turk & continuous scale & perceived\_message\_effectiveness; argument\_strength; attitude toward vaccination \\
S20 & public health messaging & message exposure with no source label & Amazon Mechanical Turk & continuous scale & perceived\_message\_effectiveness; argument\_strength; attitude toward vaccination \\
S21 & health communication / vaccine persuasion & multi-turn conversation & Prolific and CloudResearch Connect & continuous intention & Immediate HPV vaccine likelihood (0--100) \\
S22 & political persuasion & exposure to a persuasive message & Prolific & attitude/policy support & policy support for a smoking ban \\
S23 & political persuasion & exposure to a persuasive message & Prolific and CloudResearch & attitude/policy support & policy support for an assault weapons ban \\
S24 & political persuasion & exposure to a persuasive message & Prolific & attitude/policy support & policy support across four policies \\
S26 & health communication & source disclosure labels & two study pools (University study pool and Prolific) & attitude/evaluation & effects perception (EP) under moderation by negative attitudes toward AI \\
\end{longtable}

\end{landscape}

\end{document}